\documentclass[runningheads]{llncs}
\usepackage{graphicx}
\usepackage{algpseudocode}
\usepackage[ruled]{algorithm}
\usepackage{amsmath}
\usepackage{amssymb}
\usepackage{amsfonts}
\usepackage{xcolor}
\usepackage{subfig}
\usepackage[hyphens]{url}

\begin{document}
\title{A Stochastic Alternating Balance $k$-Means Algorithm for Fair Clustering}
%
%
\author{Suyun Liu \inst{1}\orcidID{0000-0002-9226-7436} \and
Luis Nunes Vicente\inst{2}\orcidID{0000-0003-1097-6384}}
\authorrunning{S. Liu and L. N. Vicente.}
%
\institute{Department of Industrial and Systems Engineering, Lehigh University, Bethlehem, PA 18015, USA. \email{sul217@lehigh.edu}.
\and Department of Industrial and Systems Engineering, Lehigh University, Bethlehem, PA 18015, USA. \email{lnv@lehigh.edu}. Support for this author was partially provided by the Centre for Mathematics of the University of
Coimbra under grant FCT/MCTES UIDB/MAT/00324/2020.}
%
\maketitle              
\begin{abstract}
In the application of data clustering to human-centric decision-making systems, such as loan applications and advertisement recommendations, the clustering outcome might discriminate against  people across different demographic groups, leading to unfairness. A natural conflict occurs between the cost of clustering (in terms of distance to cluster centers) and the balance representation of all demographic groups across the clusters, leading to a bi-objective optimization problem that is nonconvex and nonsmooth. To determine the complete trade-off between these two competing goals, we design a novel stochastic alternating balance fair $k$-means (SAfairKM) algorithm, which consists of alternating classical mini-batch $k$-means updates and group swap updates. The number of $k$-means updates and the number of swap updates essentially parameterize the weight put on optimizing each objective function. Our numerical experiments show that the proposed SAfairKM algorithm is robust and computationally efficient in constructing well-spread and high-quality Pareto fronts both on synthetic and real datasets. 

\end{abstract}

\section{Introduction}

Clustering is a fundamental task in data mining and unsupervised machine learning with the goal of partitioning data points into clusters, in such a way that data points in one cluster are very similar and data points in different clusters are quite distinct~\cite{GGan_CMa_JWu_2020}. It has become a core technique in a huge amount of application fields such as feature engineering, information retrieval, image segmentation, targeted marketing, recommendation systems, and urban planning. Data clustering problems take on many different forms, including partitioning clustering like $k$-means and $k$-median, hierarchical clustering, spectral clustering, among many others~\cite{PBerkhin_2006,GGan_CMa_JWu_2020}.  Given the increasing impact of automated decision-making systems in our society, 
there is a growing concern about algorithmic unfairness, which in the case of clustering may result in discrimination against minority groups. For instance, females may receive proportionally fewer job recommendations with high salary~\cite{ADatta_MCTschantz_ADatta_2015} due to their under-representation in the cluster of high salary recommendations. Such demographic features like gender and race are called \textit{sensitive} or \textit{protected} features, which we wish to be fair with respect to. 

\paragraph{Related work} An extensive literature work studying algorithmic fairness has been focused on developing universal fairness definitions and designing fair algorithms for supervised machine learning problems. Among the broadest representative fairness notions proposed for classification and regression tasks are \textit{disparate impact}~\cite{SBarocas_ADSelbst_2016} (also called~\textit{demographic parity}~\cite{TCalders_FKamiran_MPechenizkiy_2009}), \textit{equalized odds}~\cite{MHardt_EPrice_NSrebro_2016}, and individual fairness~\cite{CDwork_2012}, based on which the fairness notions in clustering were proposed accordingly. 
There are a number of classes of fairness definitions proposed and investigated for the clustering task~\cite{MAbbasi_ABhaskara_SVenkatasubramanian_2020,XChen_etal_2019,FChierichetti_etal_2017,MGhadiri_SSamadi_SVempala_2021,MKleindessner_PAwasthi_JMorgenstern_2020,SMahabadi_AVakilian_2020}. The most widely used fairness notion is called \textit{balance}. It was proposed by~\cite{FChierichetti_etal_2017}, and it has been extended in several subsequent works~\cite{SBera_etal_2019,LHuang_SJiang_NVishnoi_2019,MSchmidt_CSchwiegelshohn_CSohler_2018}.
As a counterpart of the disparate impact concept in fair supervised machine learning, balance essentially aims at ensuring that the representation of protected groups in each cluster preserves the global proportion of each protected group.

Depending on the stage of clustering in which the fairness requirements are imposed, the prior works on fair clustering are categorized into three families, namely pre-processing, in-processing, and post-processing. A large body of the literature work~\cite{ABackurs_etal_2019,FChierichetti_etal_2017,LHuang_SJiang_NVishnoi_2019,MSchmidt_CSchwiegelshohn_CSohler_2018} falls into the pre-processing category. The whole dataset is first decomposed into small subsets named \textit{fairlets}, where the desired balance can be guaranteed. Any resulting solution from classical clustering algorithms using the set of fairlets will then be fair. Chierichetti et al.~\cite{FChierichetti_etal_2017} focused on the case of two demographic groups and formulated explicit combinatorial problems (such as perfect matching and minimum cost flow problems) to decompose the dataset into minimal fair sets defining the fairlets. Their theoretical analysis gave strong guarantees on the quality of the fair clustering solutions for $k$-center and $k$-median problems. Following that line of work, Backurs et al.~\cite{ABackurs_etal_2019} embedded the whole dataset into a hierarchical structure tree and improved the time complexity of the fairlet decomposition step from quadratic to nearly linear time (in the dataset size).  Schmidt et al.~\cite{MSchmidt_CSchwiegelshohn_CSohler_2018} introduced the notion of fair \textit{coresets} and proposed an efficient streaming fair clustering algorithm for $k$-means.  
They introduced a near-linear time algorithm to construct coresets that helps reduce the input data size and hence speeds up any fair clustering algorithm. 
Huang et al.~\cite{LHuang_SJiang_NVishnoi_2019} further boosted the efficiency of coresets construction and made a generalization to multiple non-disjoint demographic groups for both $k$-means and $k$-median. 

On the contrary, post-processing clustering methods~\cite{SAhmadian_etal_2019,SBera_etal_2019,MKleindessner_PAwasthi_JMorgenstern_2019,CROsner_MSchmidt_2018} modify the resulting clusters from classical clustering algorithms to improve fairness. For example, Bera et al.~\cite{SBera_etal_2019} proposed a fair re-assignment problem as a linear relaxation of an integer programming model given the clustering results from any vanilla $k$-means, $k$-median, or $k$-center algorithms. They showed how to derive a $(\rho + 2)$-approximation fair clustering algorithm from any $\rho$-approximation vanilla clustering algorithm within a theoretical bound of fairness constraints violation. Moreover, their framework works for datasets with multiple and potentially overlapping demographic groups.  
Lastly, in-processing methods incorporate the fairness constraints into the clustering process~\cite{SSAbraham_SSSundaram_2019,MKleindessner_etal_2019,IMZiko_etal_2019}. Our approach falls into this category and allows for the determination of the trade-offs between clustering costs and fairness. To our knowledge, the only such other in-processing approach is the one of Ziko et al.~\cite{IMZiko_etal_2019}, where the clustering balance is approximately measured by the KL-divergence and imposed as a penalty term in the fair clustering objective function. The penalty coefficient is then used to control the trade-offs between clustering cost/fairness.

\paragraph{Our contribution} The partitioning clustering model, also referred to as the center-based clustering model, consists of selecting a certain number $K$ of centers and assigning data points to their closest centers. In this paper, we will focus on the well-known $k$-means model, and we will introduce a novel fair clustering algorithm using the balance measure. The main challenge of the fair clustering task comes from the violation of the assignment routine, which then indicates that a data point is no longer necessarily assigned to its closest cluster. The higher the balance level one wants to achieve, the more clustering cost is added to the final clustering. Hence, there exists a natural conflict between the fairness level, when measured in terms of balance, and the classical $k$-means clustering objective. 

We explicitly formulate the trade-offs between the $k$-means clustering cost and the fairness as a bi-objective optimization problem, where both objectives are written as nonconvex and nonsmooth functions of binary assignment variables defining point assignments in the clustering model (see~\eqref{biobjective_faircluster} further below). Our goal is to construct an informative approximation of the Pareto front for the proposed bi-objective fair $k$-means clustering problem, without exploring exhaustively the binary nature of the assignment variables. The most widely used method in solving general bi-objective optimization problems is the so-called weighted-sum method~\cite{SGass_TSaaty_1955}. There, one considers a set of single objective problems, formed by convex linear combinations of the two functions, and (a portion of) the Pareto front might be approximated by solving the corresponding weighted-sum problems. However, this methodology has no rigorous guarantees due to the nonconvexity of both objective functions. Also, the non-smoothness of the fairness objective poses an additional difficulty to the weighted-sum method, as one function will be smooth and the other one no. Moreover, even ignoring the nonconvexity and non-smoothness issues, the two objectives, namely the clustering cost and the clustering balance, can have significantly different magnitudes. One can hardly preselect a good set of weights corresponding to decision-makers' preferences to capture a well-spread Pareto~front.

Therefore, we were motivated to design a novel stochastic alternating balance fair $k$-means (SAfairKM) algorithm, inspired from the classical mini-batch $k$-means algorithm, which essentially consists of alternatively taking pure mini-batch $k$-means updates and swap-based balance improvement updates.  In fact, the number of $k$-means updates (denoted by $n_a$) and the number of swap updates (denoted by $n_b$) play a role similar to the weights in the weighted-sum method, parameterizing the efforts of optimizing each objective. In the pure mini-batch $k$-means updates, we focus on minimizing the clustering cost. A mini-batch of points is randomly drawn and assigned to their closest clusters, after which the set of centers are updated using mini-batch stochastic gradient descent. In the swap-based balance improvement steps, we aim at increasing the overall clustering balance. For this purpose, we propose a simple swap routine that is guaranteed to increase the overall clustering balance by swapping data points between the minimum balance cluster and a target well-balanced cluster. Similarly to the $k$-means updates, the set of centers are updated using the batch of data points selected to swap. While the $k$-means updates reproduce the stochastic gradient descent directions for the clustering cost function, the swap updates can be seen as taking steps along some increasing directions for the clustering balance objective (not necessarily the best ascent direction).

We have evaluated the performance of the proposed SAfairKM algorithm using both synthetic datasets and real datasets. 
To endow SAfairKM with the capability of constructing a Pareto front in a single run, we use a list of nondominated points updated at every iteration. The list is randomly generated at the beginning of the process. At every iteration, and for every point in the current list, we apply SAfairKM for all considered pairs of $(n_a, n_b)$. For each pair $(n_a, n_b)$, one does $n_a$ $k$-means updates and $n_b$ swap updates. At the end of each iteration, we remove from the list all dominated points (those for each there exists another one with higher clustering cost and lower clustering balance). Such a simple mechanism is also beneficial for excluding bad local optima, considering that the two objectives are nonconvex. 
We will present the full trade-offs between the two conflicting objectives for four synthetic datasets and two real datasets. A numerical comparison with the fair $k$-means algorithm proposed in~\cite{IMZiko_etal_2019} further confirms the robustness and efficiency of the proposed algorithm in constructing informative and high-quality trade-offs.


\section{The mini-batch $k$-means algorithm}
\label{minibatch_kmeans}
In the classical $k$-means problem, one aims to choose $K$ centers (representatives) and to assign a set of points to their closest centers. The $k$-means objective is the sum of the minimum (squared Euclidean) distance of all points to their corresponding centers.
Given a set of $N$ points $P = \{x_p\}_{p = 1}^N$, where $x_p$ is the non-sensitive feature vector, the goal of clustering is to assign $N$ points to $K$ clusters identified by $K$ centroids $C = [c_1, \ldots, c_K]^\top$. 
Let $[M]$ denote the set of positive integers up to $M$ for any $M \in \mathbb{N}$.
The $k$-means clustering problem is formulated as the minimization of a nonsmooth function of the set of centroids:
\begin{equation}
\label{kmean_cost_f1} 
    \min~f^{KM}_1(C) \;=\; \frac{1}{2}\sum_{p=1}^N \min_{k \in [K]} \|x_p - c_k\|^2.
\end{equation}
Since each data point is assigned to the closest cluster, the $K$ cluster centroids are implicitly dependent on the point assignments. Let $s_{p, k} \in \{0, 1\}$ be an assignment variable who takes the value $1$ if point $x_p$ is assigned to cluster $k$, and $0$ otherwise. For simplicity, we denote~$s_k$, $k \in [K]$, as an $N$-dimensional assignment vector for cluster~$k$, and $s_p, p \in [N]$, as a $K$-dimensional assignment vector for point $x_p$. Let $X \in \mathbb{R}^{N \times d}$ be the data matrix stacking~$N$~data points of dimension $d$ and~$e_N \in \mathbb{R}^N$ be an all-ones vector. Then one can compute each centroid using $c_k = X^\top s_k/e_N^\top s_k$.

In practice, Lloyd's heuristic algorithm~\cite{SLloyd_1982}, also known as the standard batch $k$-means algorithm, is the simplest and most popular $k$-means clustering algorithm, and converges to a local minimum but without worst-case guarantees~\cite{TKanungo_etal_2004,SZSelim_MAIsmail_1984}. The main idea of Lloyd's heuristic is to keep updating the $K$ cluster centroids and assigning the full batch of points to their closest centroids. 

In the standard batch $k$-means algorithm, one can compute the full gradient of the objective function~\eqref{kmean_cost_f1} with respect to $k$-th center by $\nabla_{c_k} f^{KM}_1 (C) = \sum_{x_p \in \mathcal{C}_k} (c_k - x_p)$,
where $\mathcal{C}_k, k \in [K],$ is the set of points assigned to cluster $k$. Whenever there exists a tie, namely a point that has the same distance to more than one cluster, one can randomly assign the point to any of such clusters. A full batch gradient descent algorithm would iteratively update the centroids by~$c^{t+1}_{k} - c^t_{k} = \alpha_k^t \sum_{x_p \in \mathcal{C}_k} (x_p - c_k), \forall k \in [K]$, where~$\alpha_k^t > 0$ is the step size. Let $N_k^t$ be the number of points in cluster $k$ at iteration $t$. It is known that the full batch $k$-means algorithm with~$\alpha_k^t = 1/N^t_k$ converges to a local minimum as fast as Newton's method, with a superlinear~rate~\cite{LBottou_YBengio_1995}. 

The standard batch $k$-means algorithm is proved to be slow for large datasets. Bottou and Bengio~\cite{LBottou_YBengio_1995} proposed an online stochastic gradient descent (SGD) variant that takes a gradient descent step using one sample at a time. Given a new data point $x_p$ to be assigned, a stochastic gradient descent step would look like $c^{t+1}_k = c^t_k + \alpha_k^t (x_p - c^t_k) \mbox{ if } x_p \mbox{ is assigned to cluster } k$.
While the SGD variant is computationally cheap for large datasets, it finds solutions of lower quality than the batch algorithm due to the stochasticity. The mini-batch version of the $k$-means algorithm uses a mini-batch sampling to lower stochastic noise and, in the meanwhile, speed up the convergence. 
The detailed mini-batch $k$-means is given in Algorithm~\ref{alg1}.
{\linespread{1.1}\addtocounter{algorithm}{-1}
\renewcommand{\thealgorithm}{{1}} 
\begin{algorithm}[!htb] 
\caption{Mini-batch $k$-means algorithm} 
\label{alg1} 
\begin{algorithmic}[1] 
\item \textbf{Input:} The set of points $P$ and an integer $K$.
\item \textbf{Output:} The set of centers $C = \{c_1, \ldots, c_K\}$.
\item Randomly select $K$ points as initial centers.
\item {\bf for} $t = 0, 1, 2, \ldots$ {\bf do}
\item \quad Randomly sample a batch of points $B_t$.
\item \quad {\bf for} $k = 1, \ldots, K$ {\bf do}
\item \quad \quad Identify the set of points $B_t^k \subseteq B_t$ whose closest center is $c_k$.
\item \quad \quad $N_{k} = N_{k} + |B_t^k|$.
\item \quad \quad $c_{k} = c_{k} + \frac{1}{N_{k}} \sum_{x_p \in B_t^k} (x_p - c_{k})$.
\par\vspace*{0.1cm}
\end{algorithmic}
\end{algorithm}
\vspace{-2ex}
}

\section{A new stochastic alternating balance fair $k$-means method}
\label{biobjective_fairclu}

\subsection{The bi-objective balance $k$-means formulation} 
Balance~\cite{FChierichetti_etal_2017} is the most widely used fairness measure in the literature of fair clustering. 
Consider $J$ disjoint demographic groups. Let $V_j$ represent the set of points in demographic group $j \in [J]$. Then, $v_{p, j}$ takes the value $1$ if point $x_p \in V_j$. We denote~$v_j$ as an $N$-dimensional indicator vector for the demographic group $j \in [J]$. The balance of cluster~$k$ is formally defined as $ b_k = \min_{j \neq j'} v_j^\top s_k/v_{j'}^\top s_k \leq 1, \forall k \in [K]$, which calculates the minimum ratio among different pairs of protected groups. The overall clustering balance is the minimum balance over all clusters, i.e., $b = \min_{k = 1}^K b_k$. The higher the overall balance, the fairer the clustering. 

By the definition of cluster balance given above, the balance function can be easily computed only using the assignment variables. The $k$-means objective~\eqref{kmean_cost_f1} can be rewritten as a function of the assignment variables as well. Hence, one can directly formulate the inherent trade-off between clustering cost and balance as a bi-objective optimization problem, i.e.,   
\begin{equation}
\label{biobjective_faircluster}
   \min \quad (f_1(s), -f_2(s)) \quad \mbox{s.t.} \quad \sum_{k=1}^K s_{p, k} = 1, \forall p \in [N], \quad s \in \{0, 1\}^{N \times K},
\end{equation}
where $s$ is the binary-valued assignment matrix with column vectors $s_k, k \in [K]$, and row vectors $s_p, p \in [N]$, and
\begin{equation*}
    f_1(s) \;=\; \frac{1}{N}\sum_{k=1}^K \sum_{p = 1}^N s_{p, k} \|x_p - c_k\|^2, \quad \text{with } c_k \;=\; \frac{X^\top s_k}{e_N^\top s_k} =  \frac{\sum_{p = 1}^N x_p s_{p, k}}{\sum_{p = 1}^N s_{p, k}},
\end{equation*}
\begin{equation*}
    f_2(s) \;=\; \min_{k \in [K]} \min_{\substack{j \neq j'\\ j, j' \in [J]}} \frac{v_j^\top s_k}{v_{j'}^\top s_k}.
\end{equation*} 
The two constraints in~\eqref{biobjective_faircluster} ensure that one point can only be assigned to one cluster. Note that both objectives are nonconvex functions of the binary assignment variables. 

\subsection{The stochastic alternating balance fair $k$-means method} 
We propose a novel stochastic alternating balance  fair $k$-means clustering algorithm to compute a nondominated solution on the Pareto front. We will use a simple but effective alternating update mechanism, which consists of improving \textit{either} the clustering objective \textit{or} the overall balance, by iteratively updating cluster centers and assignment variables. Specifically, every iteration of the proposed algorithm contains two sets of updates, namely pure $k$-means updates and pure swap-based balance improvement steps. The pure $k$-means updates were introduced in Section~\ref{minibatch_kmeans}, and will consist of taking a certain number of stochastic $k$-means steps. In the balance improvement steps, a certain batch of points is selected and swapped between the minimum balanced cluster and a target well-balanced cluster. 

\paragraph{Balance improvement steps}
At the current iteration, let $\mathcal{C}_l$ be the cluster with the minimum balance. Then $\mathcal{C}_l$ is the bottleneck cluster that defines the overall clustering balance. Without loss of generality, we assume that $b_l = |\mathcal{C}_l \cap  V_1|/|\mathcal{C}_l \cap V_2|$, which then implies that the pair of demographic groups $(V_1, V_2)$ forms a key to improve the balance of cluster $\mathcal{C}_l$, as well as the overall clustering balance. In terms of the assignment variables, we have
\begin{equation}
\label{bottleneck_balance}
    b \;=\; b_l \;=\; \frac{v_1^\top s_l}{v_2^\top s_l} \;=\; \frac{\sum_{p = 1}^N v_{p, 1} s_{p, l}}{\sum_{p = 1}^N v_{p, 2} s_{p, l}}.
\end{equation}
One way to determine a target well-balanced cluster $\mathcal{C}_h$ is to select it as the one with the maximum ratio between $V_1$ and $V_2$, i.e., 
\begin{equation}
\label{target_swap_cluster}
    h \;\in\; \text{argmax}_{k \in [K]}~ \left\{v_1^\top s_k/v_2^\top s_k, v_2^\top s_k/v_1^\top s_k \right\}.
\end{equation}
Another way to determine such a target cluster is to select a cluster  $\mathcal{C}_h$ that is closest to $\mathcal{C}_l$, i.e., 
\begin{equation}
    \label{target_swap_cluster_local}
    h \;\in\; \text{argmin}_{k \in [K], k \neq l}~ \|c_k - c_l\|.
\end{equation}
We call the target cluster computed by~\eqref{target_swap_cluster} a \textit{global} target and the one selected by~\eqref{target_swap_cluster_local} a \textit{local} target. Using a global target cluster makes the swap updates more efficient and stable in the sense that the target cluster is only changed when the minimum balanced cluster changes. Instead, swapping according to the local target leads to less increase in clustering costs.

To improve the overall balance, one swaps a point in cluster $\mathcal{C}_l$ belonging to $V_2$ with a point in cluster $\mathcal{C}_h$ belonging to $V_1$. Each of these swap updates will guarantee an increase in the overall balance. The detailed stochastic alternating balance fair $k$-means clustering algorithm is given in Algorithm~\ref{alg2}. At each iteration, we alternate between taking $k$-means updates using a drawn batch of points (denote the batch size by $n_a$) and ``swap'' updates using another drawn batch of points (denote the batch size by $n_b$). The generation of the two batches is independent. The choice of $n_a$ and $n_b$ influences the nondominated point obtained at the end, in terms of the weight put into each objective.

{\linespread{1.2}\addtocounter{algorithm}{-1}
\renewcommand{\thealgorithm}{{2}} 
\begin{algorithm}[tb] 
\caption{Stochastic alternating balance fair $k$-means clustering (SAfairKM) algorithm} 
\label{alg2} 
\begin{algorithmic}[1] 
\item \textbf{Input:} The set of points $P$, an integer $K$, and parameters $n_a, n_b$.
\item \textbf{Output:} The set of clustering labels $\Delta = \{\delta_1, \ldots, \delta_N\}$, where $\delta_p \in [K]$.
\item Randomly initialize labels $\{\delta_1, \ldots, \delta_N\}$ and a set of counters $\{N_1, \ldots, N_K\}$. Compute $k$-means centers $\{c_1, \ldots, c_K\}$ and balances $\{b_1, \ldots, b_K\}$ for all clusters.
\item {\bf for} $t = 1, 2, \ldots$ {\bf do}
\item \quad  Randomly sample a batch of $n_a$ points $B_t \subseteq P$ without replacement.
\item \quad {\bf for} $x_p \in B_t$ {\bf do}
\item \quad \quad Decrease the counter $N_{\delta_p} = N_{\delta_p} - 1$ for the previous clustering label. 
\item \quad \quad Identify its closest center index $i_p$. Update clustering label $\delta_p = i_p$.
\item \quad \quad Increase the counter $N_{\delta_p} = N_{\delta_p} + 1$ and center $c_{\delta_p} = c_{\delta_p} + \frac{1}{N_{\delta_p}} (x_p - c_{\delta_p})$.
\item \quad {\bf for} $r = 1, 2, \ldots, n_b$ {\bf do}
\item \quad \quad Identify $\mathcal{C}_l$, $\mathcal{C}_h$, and the pair of demographic groups $(V_1, V_2)$ according to~\eqref{bottleneck_balance} and~\eqref{target_swap_cluster_local}.
\item \quad \quad Randomly select points $x_p \in \mathcal{C}_l \cap V_{2}$ and $x_{p'} \in \mathcal{C}_h \cap V_{1}$.
\item \quad \quad Swap points: set $\delta_p = h$ and $\delta_{p'} = l$.
\item \quad \quad Update centers $c_l = c_l + \frac{1}{N_l} (x_{p'} - c_l)$ and $c_h = c_h + \frac{1}{N_h} (x_p - c_h)$.
\item \quad \quad Update balance for clusters $\mathcal{C}_l$ and $\mathcal{C}_h$.
\par\vspace*{0.1cm}
\end{algorithmic}
\end{algorithm}
}

Instead of randomly selecting points to swap in line 11 of Algorithm~\ref{alg2}, in our experiments we have used a more accurate swap strategy by increasing the batch size. Basically, we randomly sample a batch of points from $\mathcal{C}_l \cap V_2$ (resp. $\mathcal{C}_h \cap V_1$) and select $x_p$ (resp. $x_p'$) as the one closest to $\mathcal{C}_h$ (resp. $\mathcal{C}_l$). The batch size could be increased as the algorithm proceeds. Our numerical experiments show that the combination of local target clusters and the increasingly accurate swap strategy result in better numerical performance.

One could have converted the bi-objective optimization problem~\eqref{biobjective_faircluster} into a weighted-sum function using the weights associated with the decision-maker's preference. However, optimizing such a weighted-sum function hardly reflects the desired trade-off due to significantly different magnitudes of the two objectives. Moreover, the existing $k$-means algorithm frameworks, including Lloyd's heuristic algorithm, are not capable of directly handling the weighted-sum objective function. In our proposed SAfairKM algorithm, the pair $(n_a, n_b)$ plays a role similar to the weights in the weighted-sum method. 

\section{Numerical experiments}
\label{numerical_exp}

\subsection{Pareto front SAfairKM algorithm}
\label{pf_SAfairKM}
In our implementation\footnote{Our implementation code is available at~\url{https://github.com/sul217/SAfairKM}. All the experiments were conducted on a MacBook Pro Intel Core i5 processor.}, to obtain a well-spread Pareto front, we frame the SAfairKM algorithm into a Pareto front version using a list updating mechanism. See Algorithm~\ref{alg:PF_SAfairKM} for a detailed description. In the initialization phase, we specify a sequence of pairs of the number of $k$-means updates and swap updates $\mathcal{W} = \{(n_a, n_b): n_a + n_b = n_{\text{total}}, n_a, n_b \in \mathbb{N}_0\}$, and we generate a list of random initial clustering labels $\mathcal{L}_0$. Then we run Algorithm~\ref{alg2} for a certain number of iterations ($q=1$ in our experiments) parallelly for each label in the current list $\mathcal{L}_t$, resulting in a new list of clustering labels $\mathcal{L}_{t+1}$. At the end of each iteration, the list is cleaned up by removing all the dominated points from $\mathcal{L}_{t+1}$. Using this algorithm, the list of nondominated points is refined towards the true real Pareto front. The process can be terminated when either the number of nondominated points is greater than a certain budget ($1500$ in our experiments) or when the total number of iterations exceeds a certain limit (depending on the size of the dataset).

\vspace{-3ex}
{\linespread{1}\addtocounter{algorithm}{-1}
\renewcommand{\thealgorithm}{{3}} 
\begin{algorithm}[H] 
\caption{Pareto-Front SAfairKM Algorithm}
\label{alg:PF_SAfairKM}
\begin{algorithmic}[1] 
\par\vspace*{0.1cm}
\item
Generate a list of starting labels $\mathcal{L}_0$. Select parameter $q \in \mathbb{N}$ and a sequence of pairs $\mathcal{W} = \{(n_a, n_b): n_a + n_b = n_{\text{total}}, n_a, n_b \in \mathbb{N}_0\}$.
\item {\bf for} $t=0,1, \ldots$ {\bf do}
\item \quad\quad Set $\mathcal{L}_{t+1} = \mathcal{L}_t$.
\item \quad\quad {\bf for} each clustering label $\Delta$ in the list $\mathcal{L}_{k+1}$ {\bf do}
\item \quad\quad\quad\quad {\bf for} $(n_a, n_b) \in \mathcal{W}$ {\bf do}
\item \quad \quad\quad\quad\quad Apply $q$ iterations of Algorithm~\ref{alg2} starting from~$\Delta$ using the parameters $(n_a, n_b)$.
 \item \quad \quad\quad\quad\quad Add the final output label to the list $\mathcal{L}_{t+1}$.
\item \quad\quad Remove all the dominated points from $\mathcal{L}_{t+1}$:
{\bf for} each label~$\Delta$ in the list $\mathcal{L}_{t+1}$ {\bf do}
\item \quad\quad If $\exists~\Delta' \in \mathcal{L}_{t+1}$ such that $ f_1(\Delta') < f_1(\Delta)$ and $f_2(\Delta') > f_2(\Delta)$ hold, remove~$\Delta$.
\par\vspace*{0.1cm}
\end{algorithmic}
\end{algorithm}
\vspace{-3ex}
}

To the best of our knowledge, the only approach in the literature providing a mechanism of controlling trade-offs between the two conflicting objectives was suggested by~\cite{IMZiko_etal_2019} and briefly described in Appendix~\ref{vfairkm}. Their approach (here called VfairKM) consists of solving~\eqref{kl_fairclu} for different penalty coefficients~$\mu$, resulting in a set of solutions from which we then remove dominated solutions to obtain an approximated Pareto front. To ensure a fair comparison, we select a set of penalty coefficients evenly from $0$ to an upper bound $\mu_{\max}$, which is determined by pre-experiments such that the corresponding fairness error is less than $0.01$ or no longer possibly decreased when further increasing its value. In some cases, we found that VfairKM is not able to produce a fairer clustering outcome when the penalty coefficient is greater than $\mu_{\max}$ due to numerical instability. 

In addition, we compare the Pareto fronts computed by SAfairKM with the fair $k$-means solution obtained from the postprocessing
fair assignment approach proposed in~\cite{SBera_etal_2019} (marked as FairAssign). In~\cite{SBera_etal_2019}, the fairest clustering solution is computed by first using a standard clustering algorithm and then applying the so-called fair assignment procedure. Such a fair assignment procedure consists of solving a linear programming relaxation of an integer programming problem, followed by an iterative rounding procedure to satisfy the bound constraints $\beta_j \leq |\mathcal{C}_k \cap V_j|/|\mathcal{C}_k| \leq \gamma_j, \forall j \in [J], k \in [K]$, where $\beta_j \in [0, 1]$ and $\gamma_j \in [0, 1]$ are lower and upper fairness bounds respectively. In our case, however, and in order to get the fairest solution, we set $\beta_j = \gamma_j = |V_j|/N$ which is exactly the proportion of demographic group $j$ in the input dataset.
Finally, we also present as benchmarks the $k$-means solutions obtained by both the state-of-the-art Lloyd’s algorithm (denoted as VanillaKM) and the mini-batch $k$-means algorithm (denoted as MinibatchKM). Both VanillaKM and MinibatchKM were equipped with the well-known $k$-means++ initialization~\cite{DArthur_SVassilvitskii_2007}.

\subsection{Numerical results}
\label{exp_res}
\paragraph{Trade-offs for synthetic datasets}
We randomly generated four synthetic datasets from Gaussian distributions, and their demographic compositions are given in Figure~\ref{syn_demo_dist} of Appendix~\ref{more_synDS_results}. Each synthetic dataset has $400$ data points in the $\mathbb{R}^2$ space and two demographic groups ($J = 2$) marked by black/circle and purple/triangle. 

\begin{figure}[!htb]
\setcounter{subfigure}{0}
\vspace{-2ex}
    \centering
    \subfloat[{\small Syn\_unequal\_ds2 ($K=2$)}.]{\includegraphics[width = 0.34\linewidth]{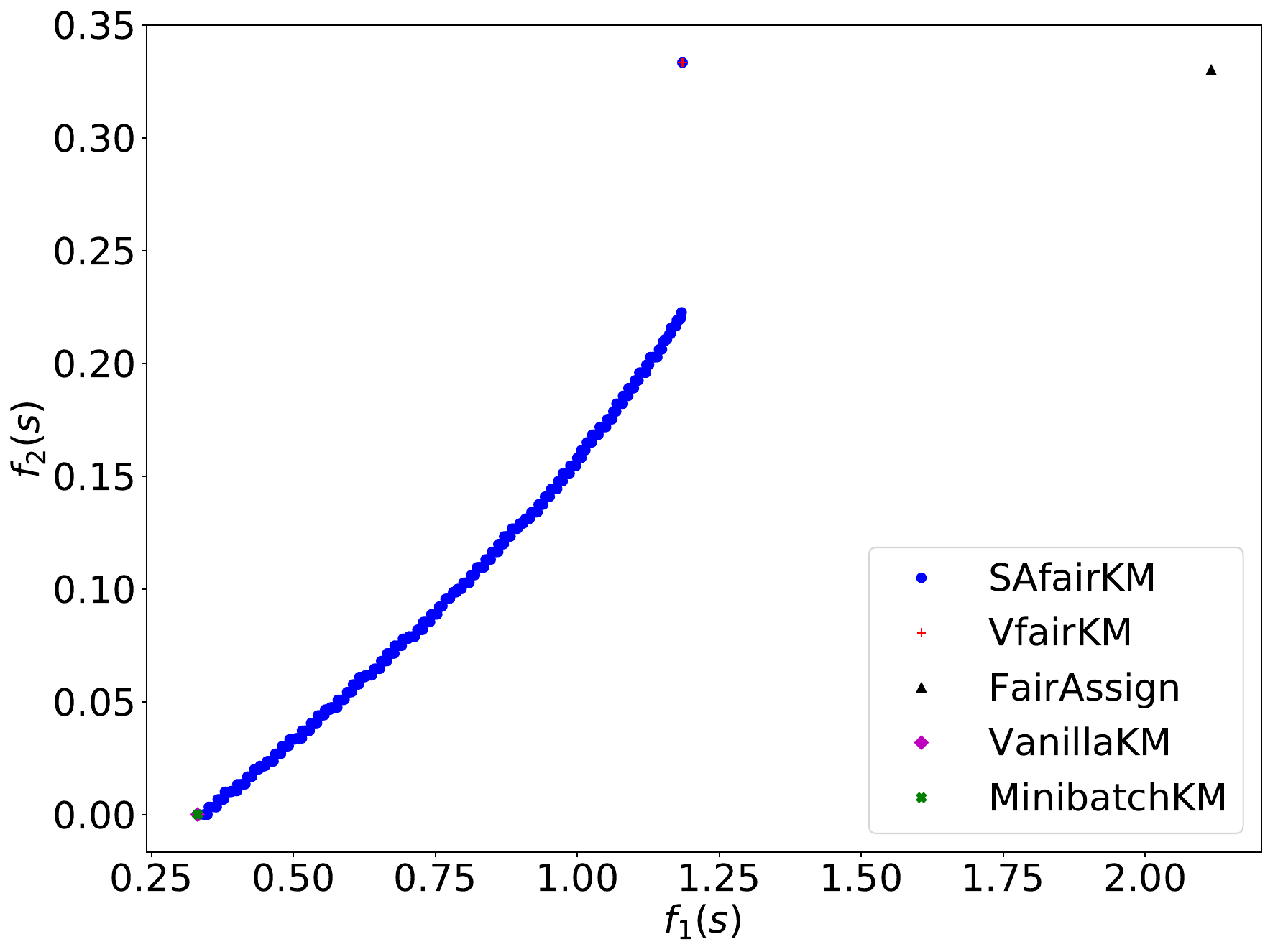}}
    \subfloat[Adult ($K = 10$).]{\includegraphics[width = 0.34\linewidth]{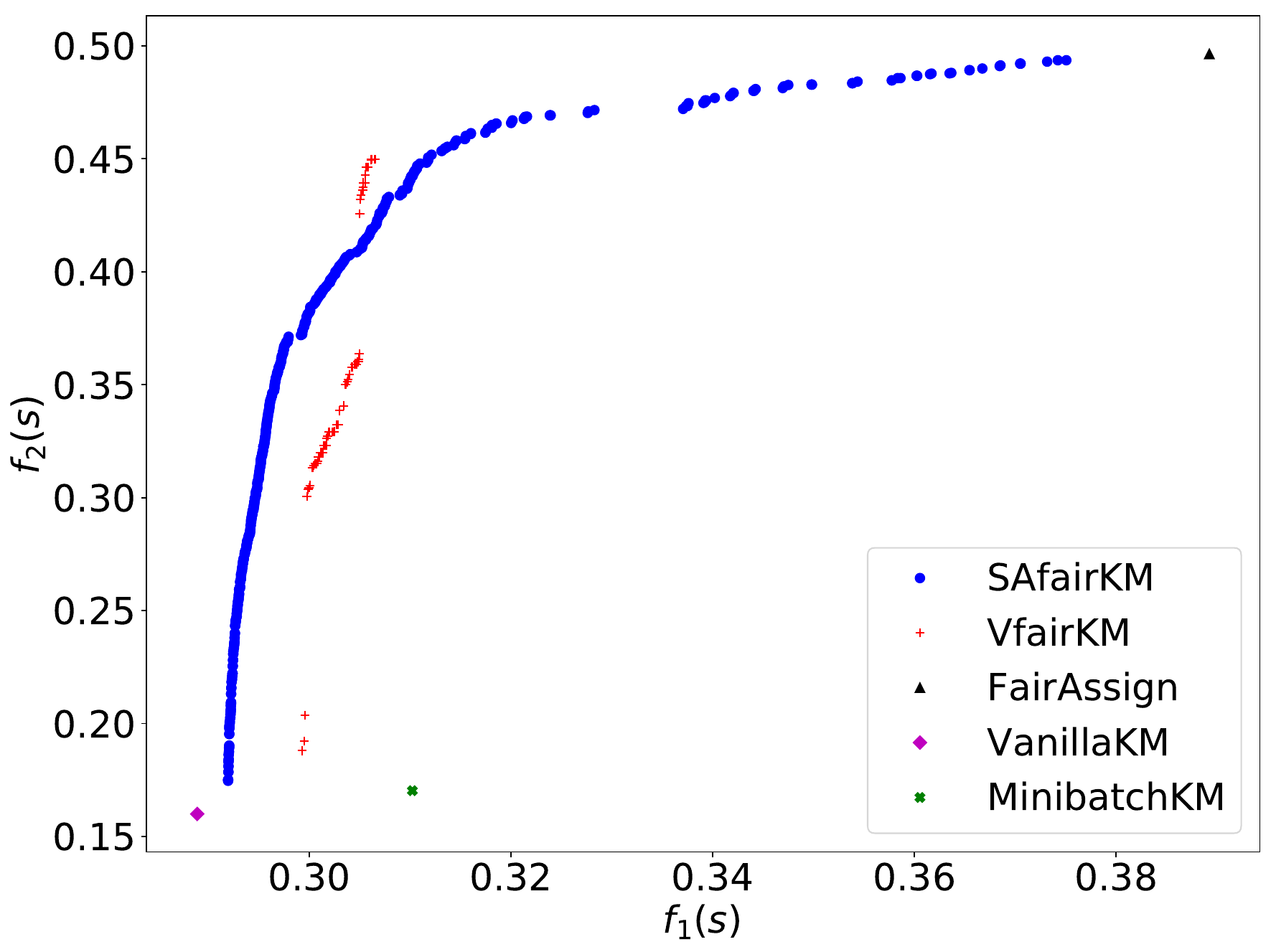}}
    \subfloat[Bank ($K = 10$).]{\includegraphics[width = 0.34\linewidth]{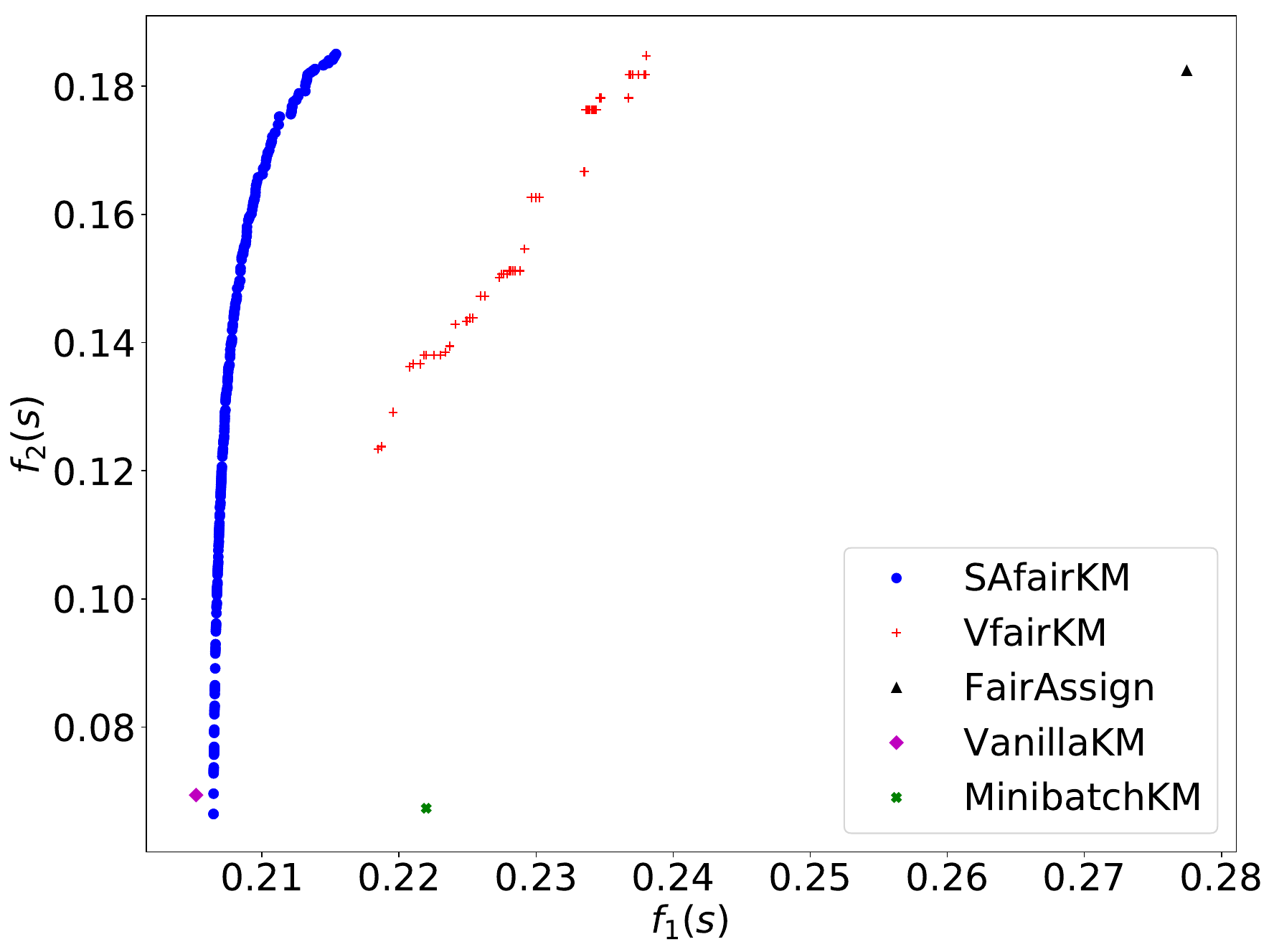}} 
    \caption{Pareto fronts: SAfairKM: $400$ iterations for Syn\_unequal\_ds2, $2500$ iterations for Adult, and $8000$ iterations for Bank, $30$ starting labels, and $4$ pairs of $(n_a, n_b)$; VfairKM: $\mu_{\max} = 0$ for Syn\_unequal\_ds2, $\mu_{\max} = 3260$ for Adult, and $\mu_{\max} = 2440$ for Bank.\label{realDs_pf}}
    \vspace{-3ex}
\end{figure}

Using the list update mechanism (described by Algorithm~\ref{alg:PF_SAfairKM}), we are able to obtain a well-spread Pareto front with comparable quality for each of the synthetic datasets. Recall that we are minimizing the clustering cost and maximizing the clustering balance. The closer the Pareto front is to the upper left corner, the higher its quality. 
In particular, Figure~\ref{realDs_pf} (a) gives the approximated Pareto front for the Syn\_unequal\_ds2 dataset with $K = 2$, which confirms the natural conflict between the clustering cost and the clustering balance. 
One can see that the VfairKM algorithm is not able to output any trade-off information as it always finds the fairest solution regardless of the value of $\mu$. Due to the special composition of this dataset, the Pareto front generated by SAfairKM is disconnected (the point around $(1.25, 0.35)$ is both VfairKM and SAfairKM). Results for the other three synthetic datasets are given by Figures~\ref{synDs1_equal}-\ref{synDs2_unequal} in Appendix~\ref{more_synDS_results}. For all the synthetic datasets, the left end point on the Pareto front given by SAfairKM is consistent with the solution of VanillaKM. On the right end of the Pareto fronts, the fair solution given by FairAssign is dominated by the fairest solution identified by our approach.

\paragraph{Trade-offs for real datasets} Two real datasets \textit{Adult}~\cite{RKohavi_1996} and \textit{Bank}~\cite{SMoro_PCortez_PRita_2014} are taken from the UCI machine learning repository~\cite{DDua_and_CGraff_2017}. The \textit{Adult} dataset contains $32,561$ samples. Each instance is characterized by $12$ nonsensitive features (including age, education, hours-per-week, capital-gain, and capital-loss, etc.). For the clustering purpose, only five numerical features among the $12$ features are kept. The demographic proportion of the \textit{Adult} dataset is $[0.67, 0.33]$ in terms of gender ($J = 2$), which corresponds to a dataset balance of $0.49$. 
The \textit{Bank} dataset contains $41,108$ data samples. 
Six nonsensitive numerical features (age, duration, number of contacts performed, consumer price index, number of employees, and daily indicator) are selected for the clustering task. Its demographic composition in terms of marital status ($J = 3$) is $[0.11, 0.28, 0.61]$, and hence the best clustering balance one can achieve is $0.185$.  

For the purpose of a faster comparison, we randomly select a subsample of size $5000$ from the original datasets and set the number of clusters to $K = 10$. The resulting solutions from the five algorithms are given in Figure~\ref{realDs_pf} (b)-(c). For both datasets, SAfairKM is able to produce more spread-out Pareto fronts which capture a larger range of balance, and hence provide more complete trade-offs between the two conflicting goals.  In terms of Pareto front quality (meaning dominance of one over the other), SAfairKM also performs better than VfairKM. In fact, we can see from Figure~\ref{realDs_pf} (b)-(c) that the Pareto fronts generated by SAfairKM dominate most of the solutions given by VfairKM and FairAssign. Also, the left end of the Pareto front generated by SAfairKM is much closer to the solution given by~VanillaKM than~VfairKM.
The Pareto fronts corresponding to $K = 5$ are also given in Figure~\ref{realDS_k5_seed1} of Appendix~\ref{more_synDS_results}. Overall, SAfairKM results in a Pareto front of higher spread and slightly lower quality than VfairKM for the Adult dataset, while the Pareto front output from SAfairKM has better spread and higher quality for the Bank dataset.
\paragraph{Performance in terms of spread and quality of Pareto fronts} SAfairKM is able to generate more spread-out and higher-quality Pareto fronts regardless of the data distribution (see the trade-off results for the four synthetic datasets). The robustness partially comes from the list update mechanism which establishes a connection among parallel runs starting from different initial points and pairs $(n_a, n_b)$, and thus helps escape from bad local optima.

\vspace{-3ex}
\begin{table}[!htb]
\centering
 \caption{Average CPU times per nondominated solution.} \label{CPU_time}
\begin{tabular}{llllll}
\hline
Dataset & SAfairKM & VfairKM & Dataset & SAfairKM & VfairKM\\
\hline
Syn\_equal\_ds1  & $0.80$ & $1.06$  & Adult ($K = 10$) & $18.52$ & $40.43$ \\ 
    Syn\_unequal\_ds1 &  $0.81$ &  $0.98$ & Bank ($K = 10$) & $59.12$ & $76.29$ \\ 
    Syn\_equal\_ds2 & $0.70$ & $1.29$ & Adult ($K = 5$) & $14.88$ & $15.08$  \\ 
    Syn\_unequal\_ds2 &  $0.80$ &  $0.10$ & Bank ($K = 5$) & $11.97$ &  $50.31$ \\
\hline
\end{tabular}
\end{table}
\vspace{-3ex}

\paragraph{Performance in terms of computational time} Since the two algorithms (SAfairKM and VfairKM) generally produce Pareto fronts of different cardinalities, we evaluate their computational efforts by the average CPU time spent per computed nondominated solution (see Table~\ref{CPU_time}). Our algorithm was shown to be clearly more computationally efficient than VfairKM.

\section{Concluding remarks}
\label{sec:conclusions}
We have investigated the natural conflict between the $k$-means clustering cost and the clustering balance from the perspective of bi-objective optimization, for which we designed a novel stochastic alternating algorithm (SAfairKM). A Pareto front version of~SAfairKM has efficiently computed well-spread and high-quality trade-offs, when compared to an existing approach based on a penalization of fairness. 

Note that a balance improvement routine for the SAfairKM algorithm could be derived to handle more than one demographic group. One might formulate a multi-objective problem with the clustering cost being one objective and the balance corresponding to each protected attribute (e.g., race and gender) written as separate objectives. The balance measured using each attribute can be improved via alternating swap updates with respect to each balance objective.

\appendix

\section{Description of an existing approach for comparison} 
\label{vfairkm}

The authors in~\cite{IMZiko_etal_2019} considered the fairness error computed by the Kullback-Leibler (KL)-divergence, and added it as a penalized term to the classical clustering objective. When using the $k$-means clustering cost, the resulting problem takes the form: 
\begin{equation}
\label{kl_fairclu}
     \min~ f_1(s) + \mu \displaystyle \sum_{k=1}^N \mathcal{D}_{KL}(U\|\mathbb{P}_k) \quad \text{s.t. } \sum_{k=1}^K s_{p, k} = 1, \forall p \in [N],
\end{equation}
where $\mathcal{D}_{KL}$ is the KL divergence between the desired demographic proportion $U = [u_j, j \in [J]]$ (usually specified by the demographic composition of the whole dataset) and the marginal probability $\mathbb{P}_k = [\mathbb{P}(j|k) = s_k^\top v_j/{e_N}^\top s_k, j \in [J]]$. 
The penalty coefficient $\mu$ associated with the fairness error is the tool to control the trade-offs between the clustering cost and the clustering balance. To solve problem~\eqref{kl_fairclu} for a fixed $\mu \geq 0$, the authors in~\cite{IMZiko_etal_2019} have developed an optimization scheme based on a concave-convex decomposition of the fairness term. 

\section{More numerical results}
\label{more_synDS_results}
\textcolor{white}{fill}

\begin{figure}[H]
\vspace{-5ex}
        \centering
        \subfloat[Syn\_equal\_ds1: $|V_1|:|V_2| = 1:1$.]{\includegraphics[width = 0.25\linewidth]{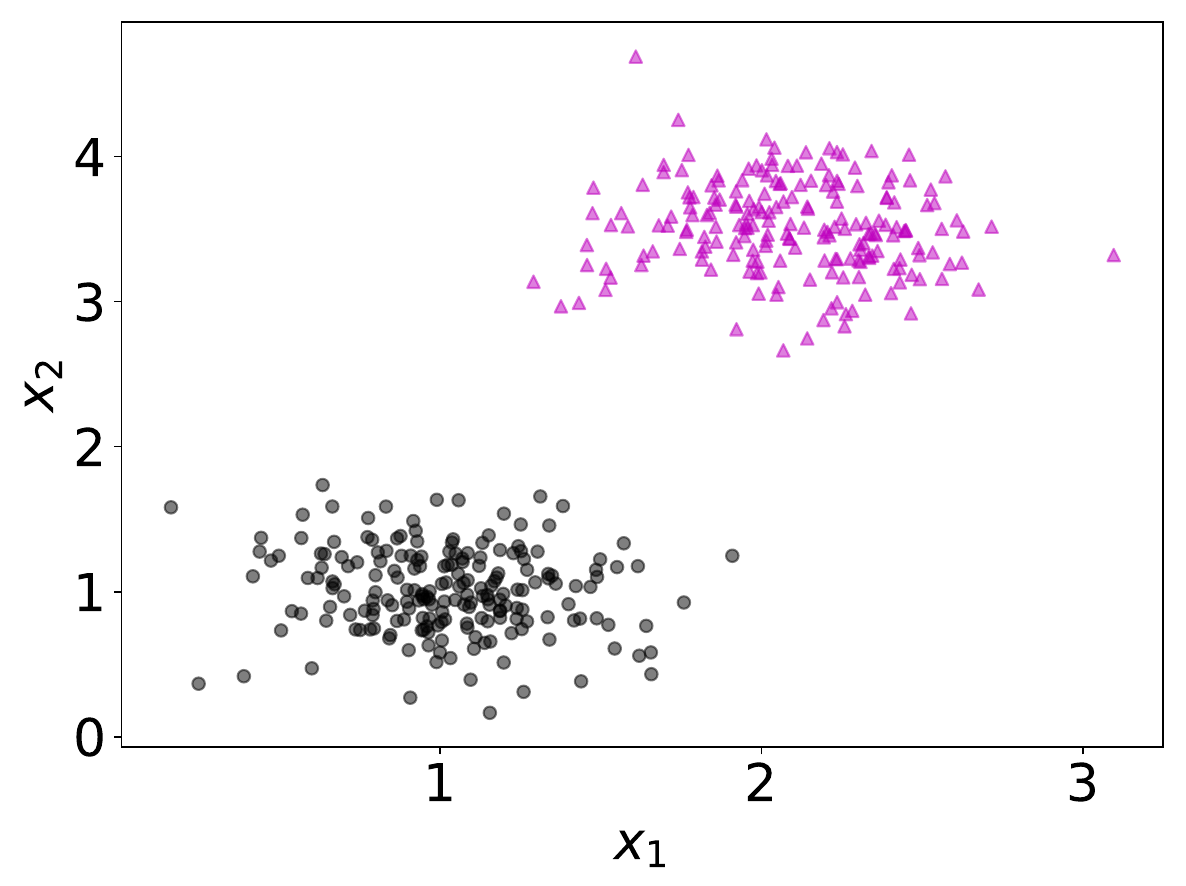}}
        \subfloat[Syn\_unequal\_ds1: $|V_1|:|V_2| = 1:3$.]{\includegraphics[width = 0.25\linewidth]{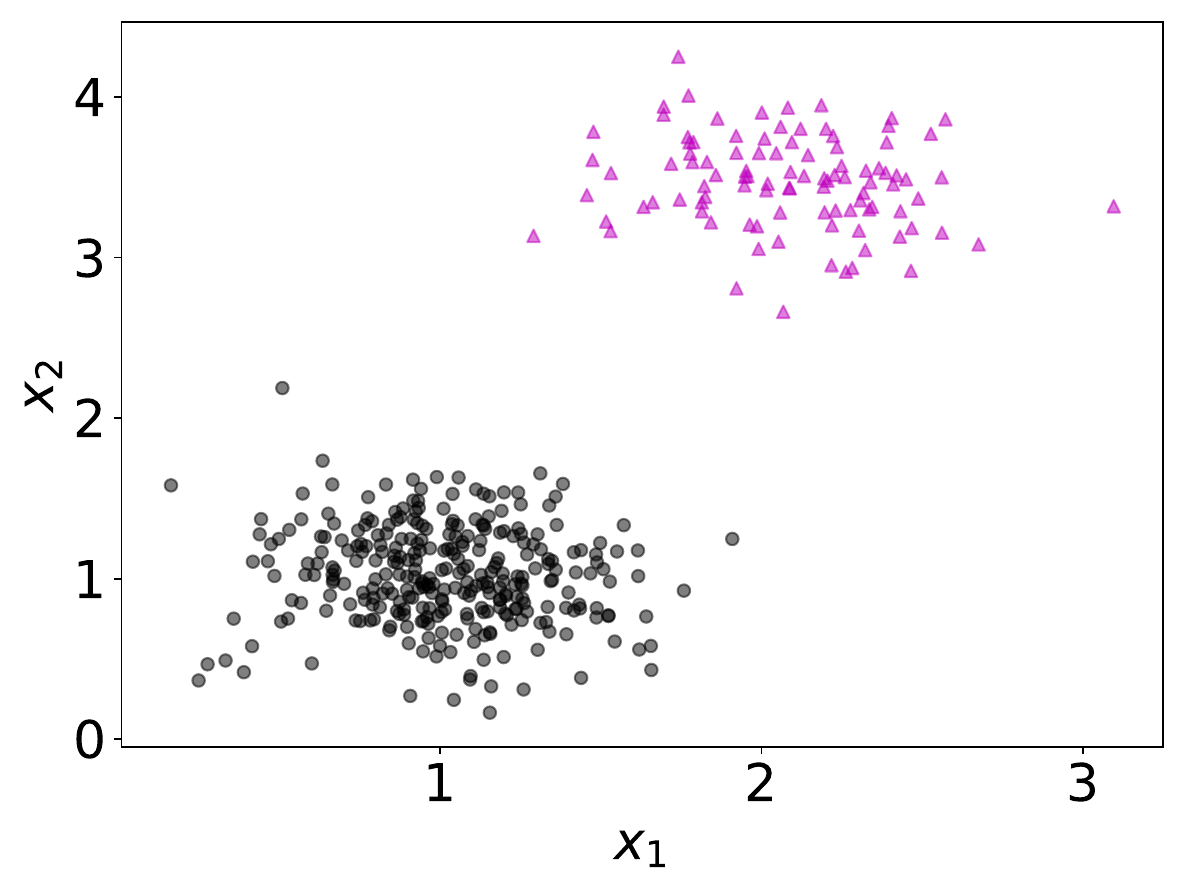}}
        \subfloat[Syn\_equal\_ds2: $|V_1|:|V_2| = 1:1$.]{\includegraphics[width = 0.25\linewidth]{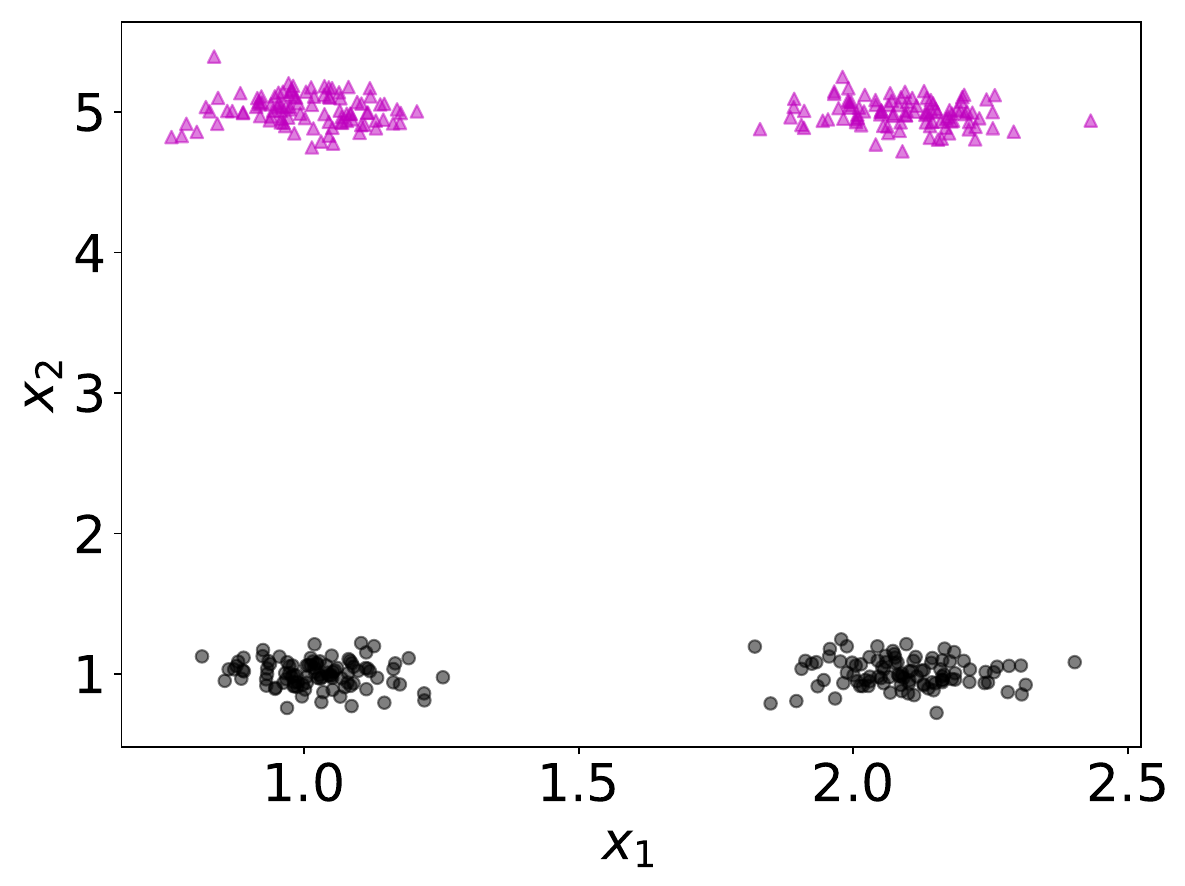}}
        \subfloat[Syn\_unequal\_ds2: $|V_1|:|V_2| = 1:3$.]{\includegraphics[width = 0.25\linewidth]{ 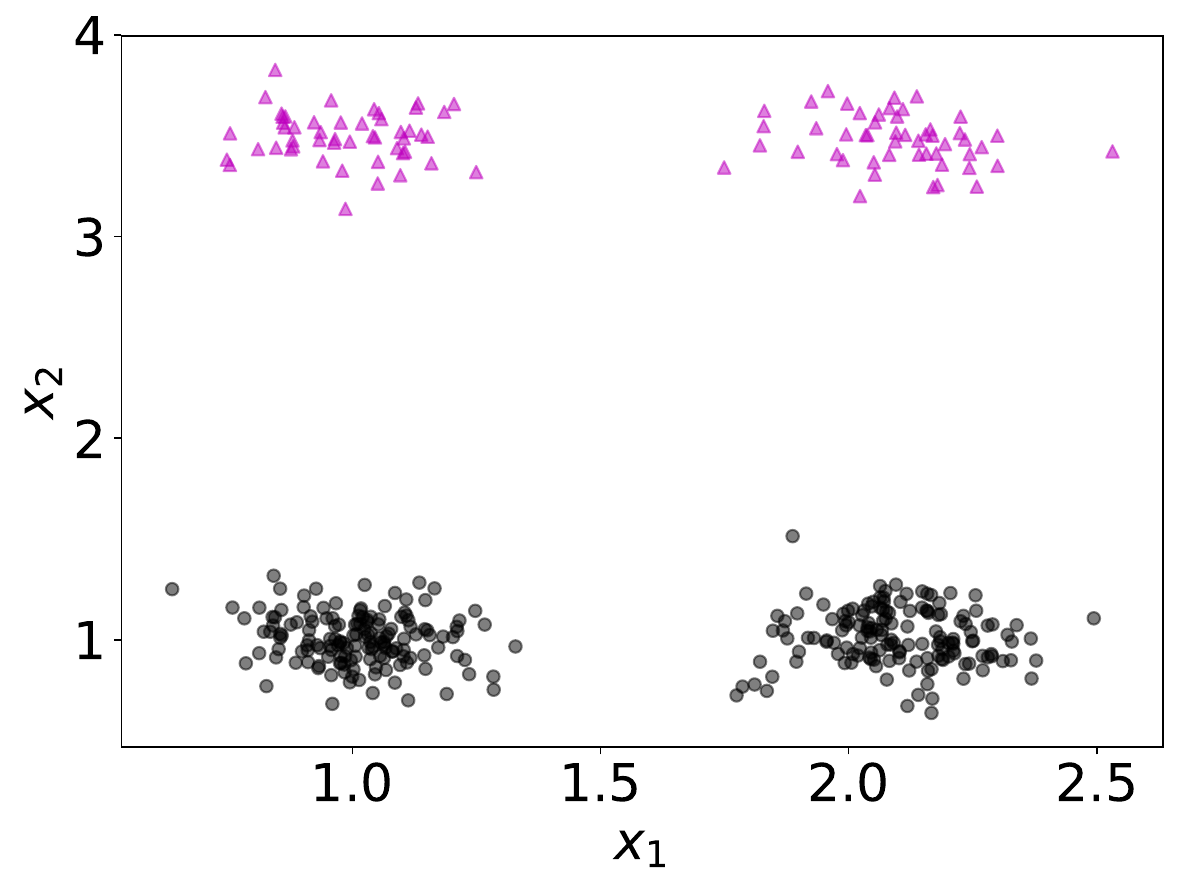}}
        \vspace{-1ex}
        \caption{Demographic composition of four synthetic datasets.\label{syn_demo_dist}}
\vspace{-4ex}
\end{figure}


\begin{figure}[!htb]
\setcounter{subfigure}{0}
    \centering
    \subfloat[Pareto front.]{\includegraphics[width = 0.25\linewidth]{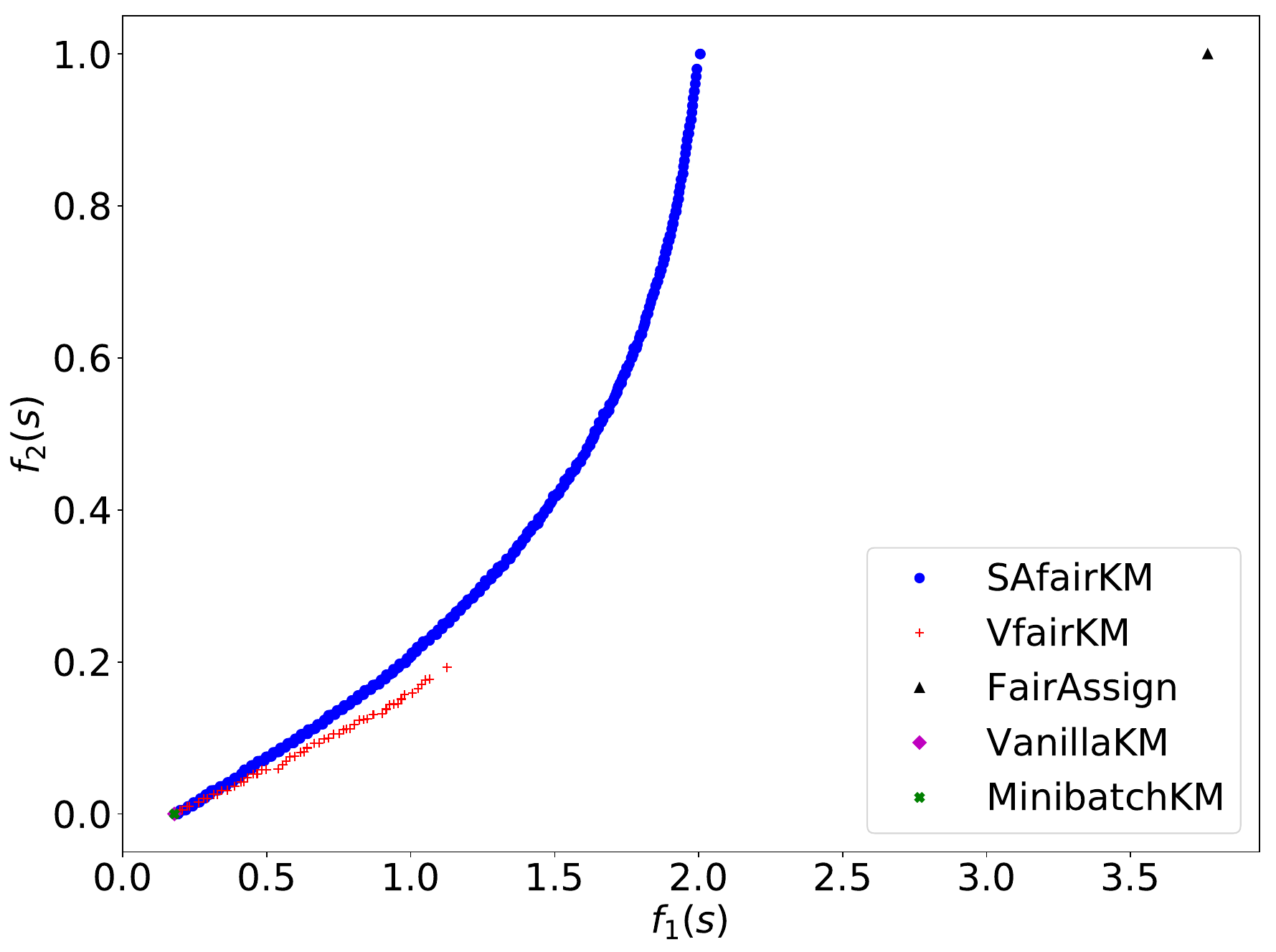}}
    \subfloat[Clust.~balance $b = 0$.]{\includegraphics[width = 0.25\linewidth]{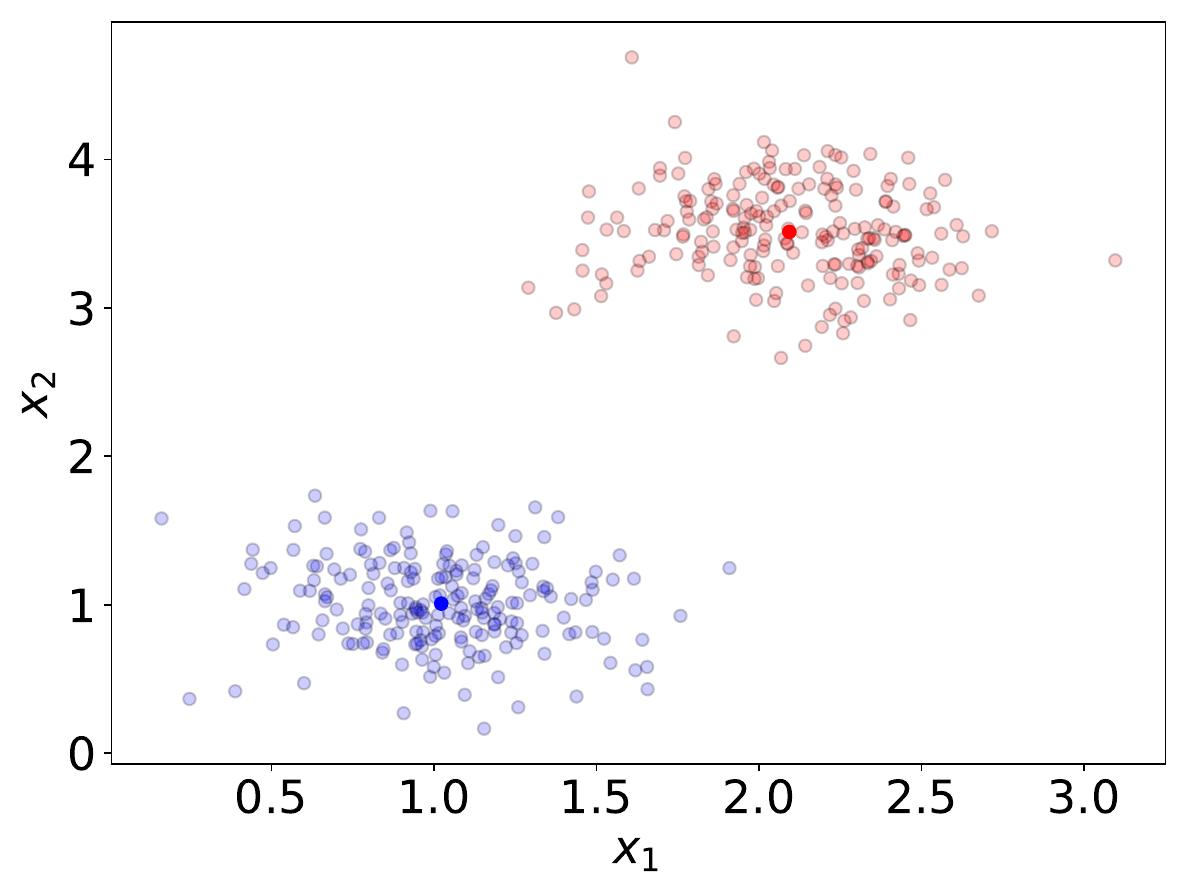}}
    \subfloat[Clust.~balance $b = 0.5$.]{\includegraphics[width = 0.25\linewidth]{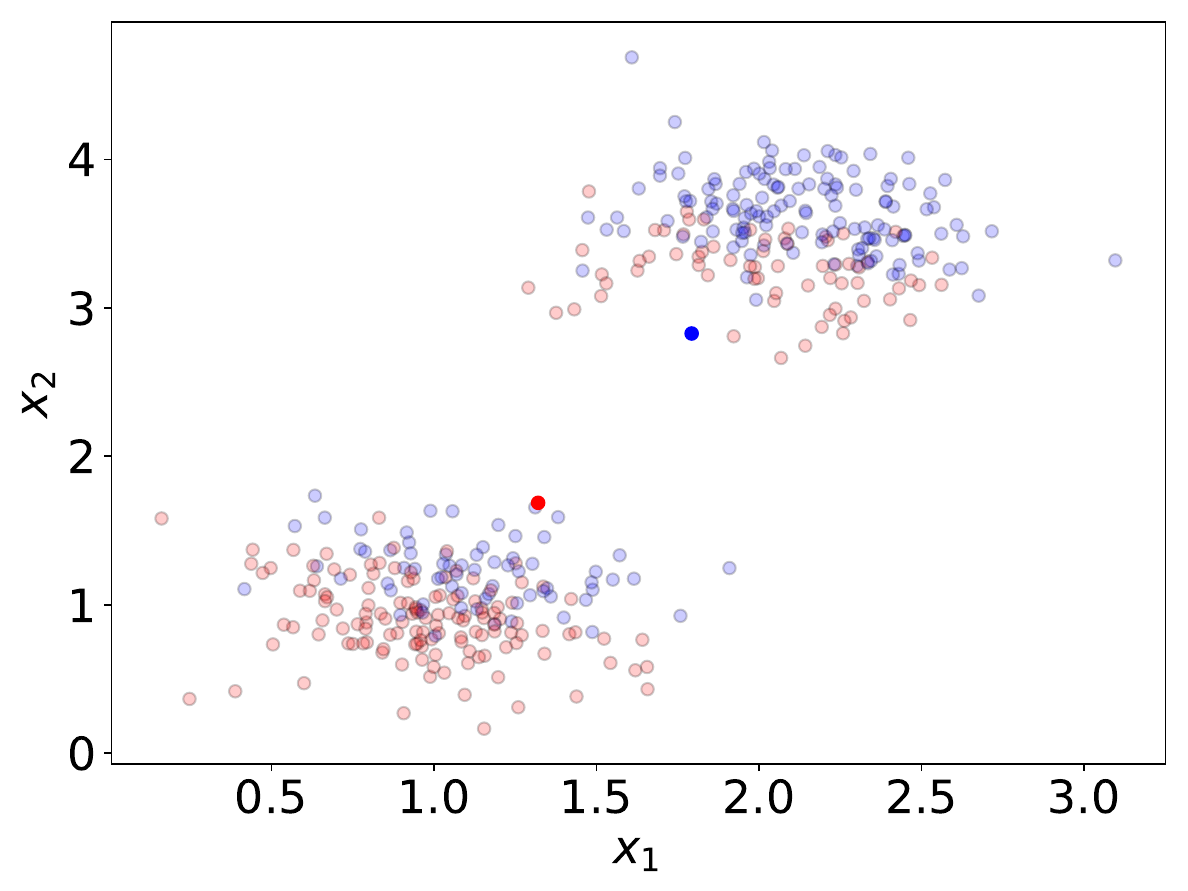}}
    \subfloat[Clust.~balance $b = 1$.]{\includegraphics[width = 0.25\linewidth]{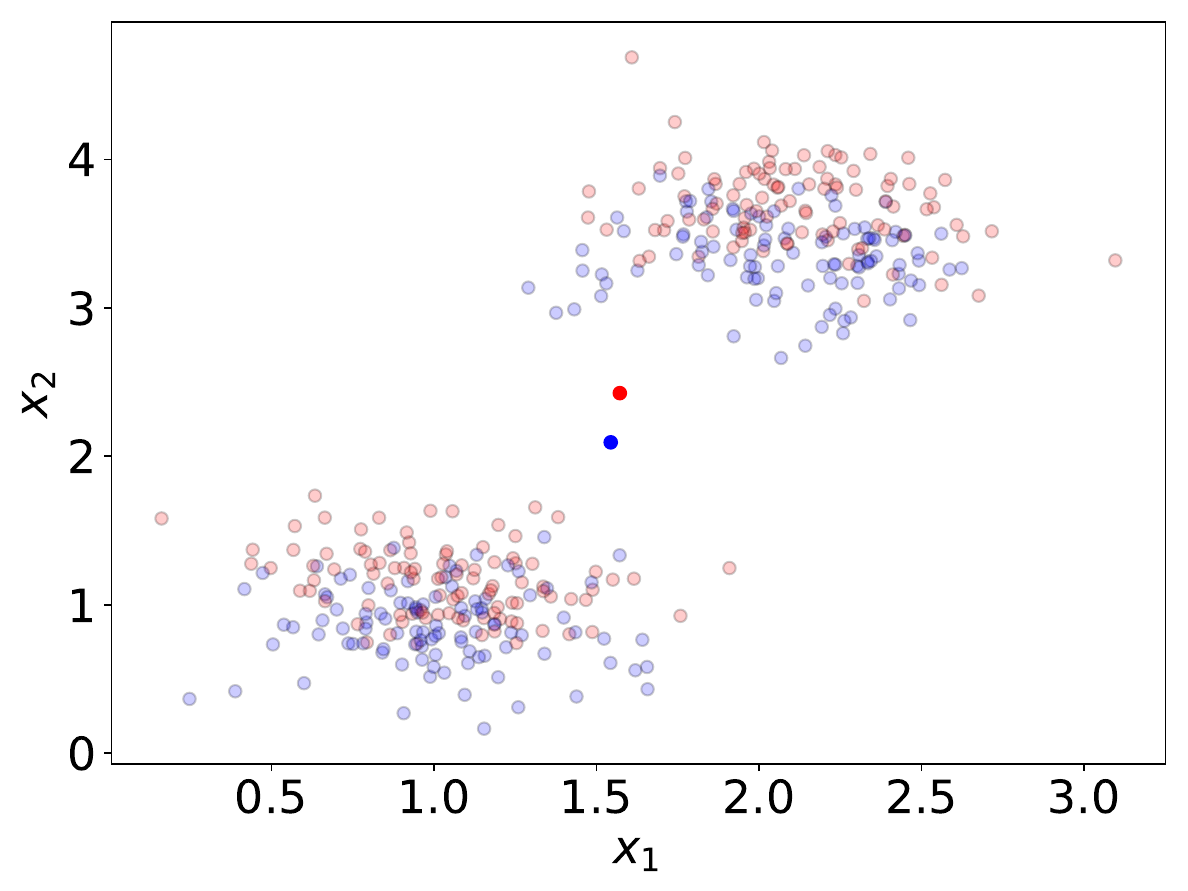}} 
    \vspace{-2ex}
    \caption{Syn\_equal\_ds1 data: SAfairKM: $400$ iterations, $10$ starting labels, and $3$ pairs of $(n_a, n_b)$; VfairKM: $\mu_{\max} = 202$. \label{synDs1_equal}}
\vspace{-2ex}
\end{figure}


\begin{figure}[!htb]
\setcounter{subfigure}{0}
    \centering
    \subfloat[Pareto front.]{\includegraphics[width = 0.25\linewidth]{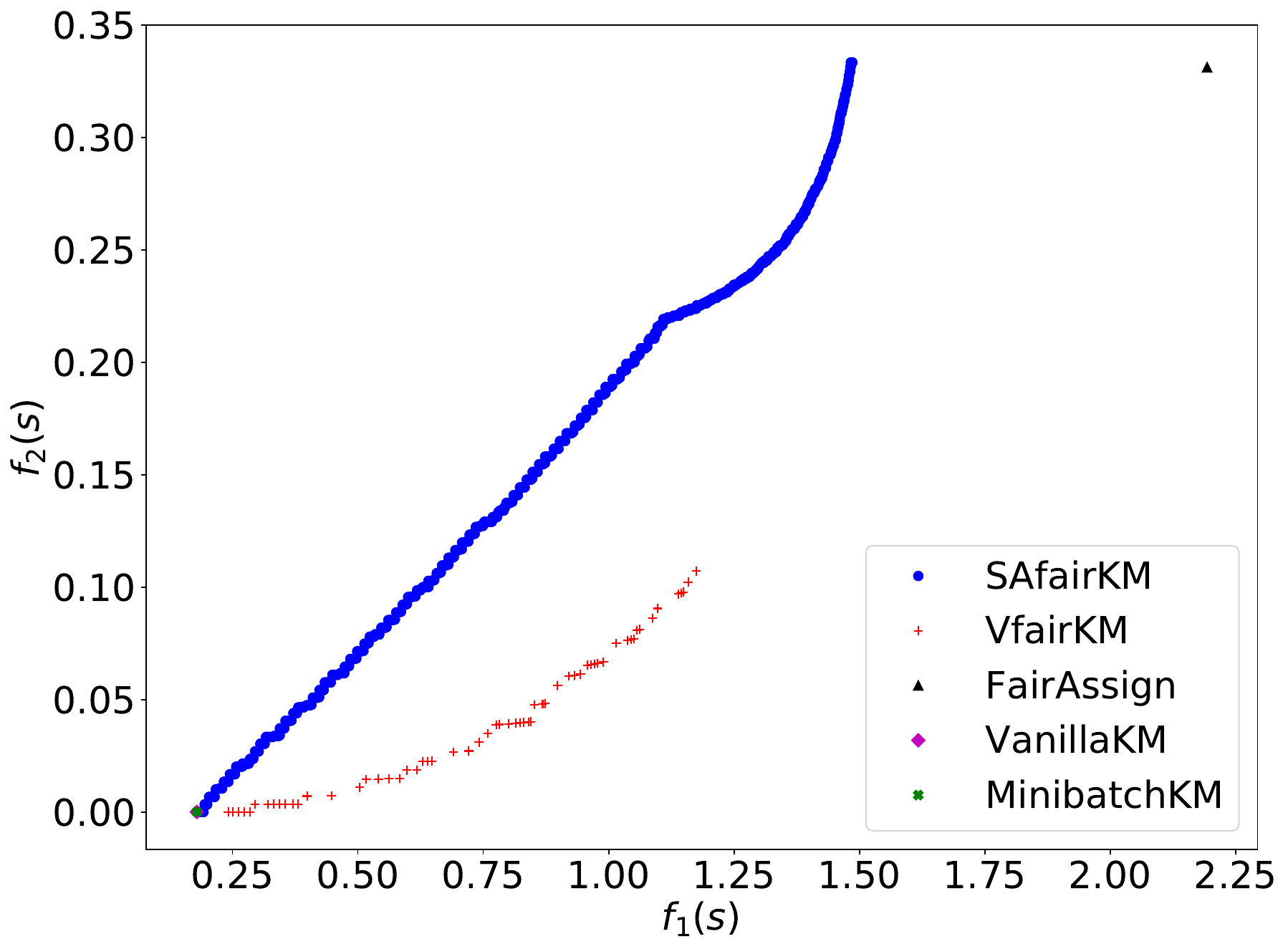}}
    \subfloat[Clust.~balance $b = 0$.]{\includegraphics[width = 0.25\linewidth]{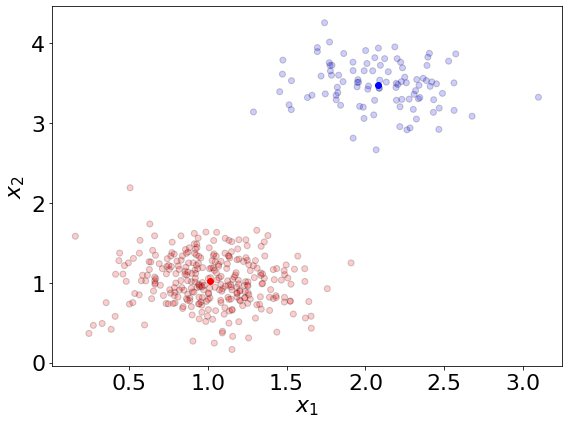}}
    \subfloat[Clust.~balance $b = 0.16$.]{\includegraphics[width = 0.25\linewidth]{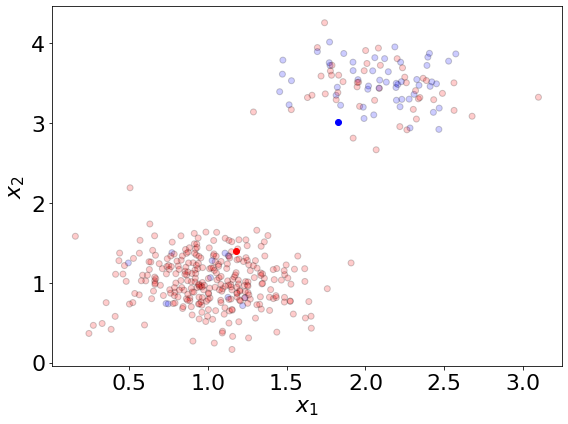}}
    \subfloat[Clust.~balance $b = 0.33$.]{\includegraphics[width = 0.25\linewidth]{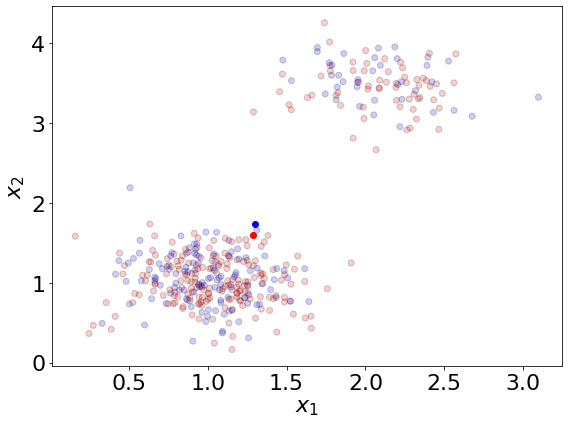}}
    \vspace{-2ex}
    \caption{Syn\_unequal\_ds1 data: SAfairKM: $400$ iterations, $10$ starting labels, and $3$ pairs of $(n_a, n_b)$; VfairKM: $\mu_{\max} = 223$.\label{synDs1_unequal}}
\vspace{-2ex}
\end{figure}

\begin{figure}[!htb]
\setcounter{subfigure}{0}
    \centering
    \subfloat[Pareto front.]{\includegraphics[width = 0.25\linewidth]{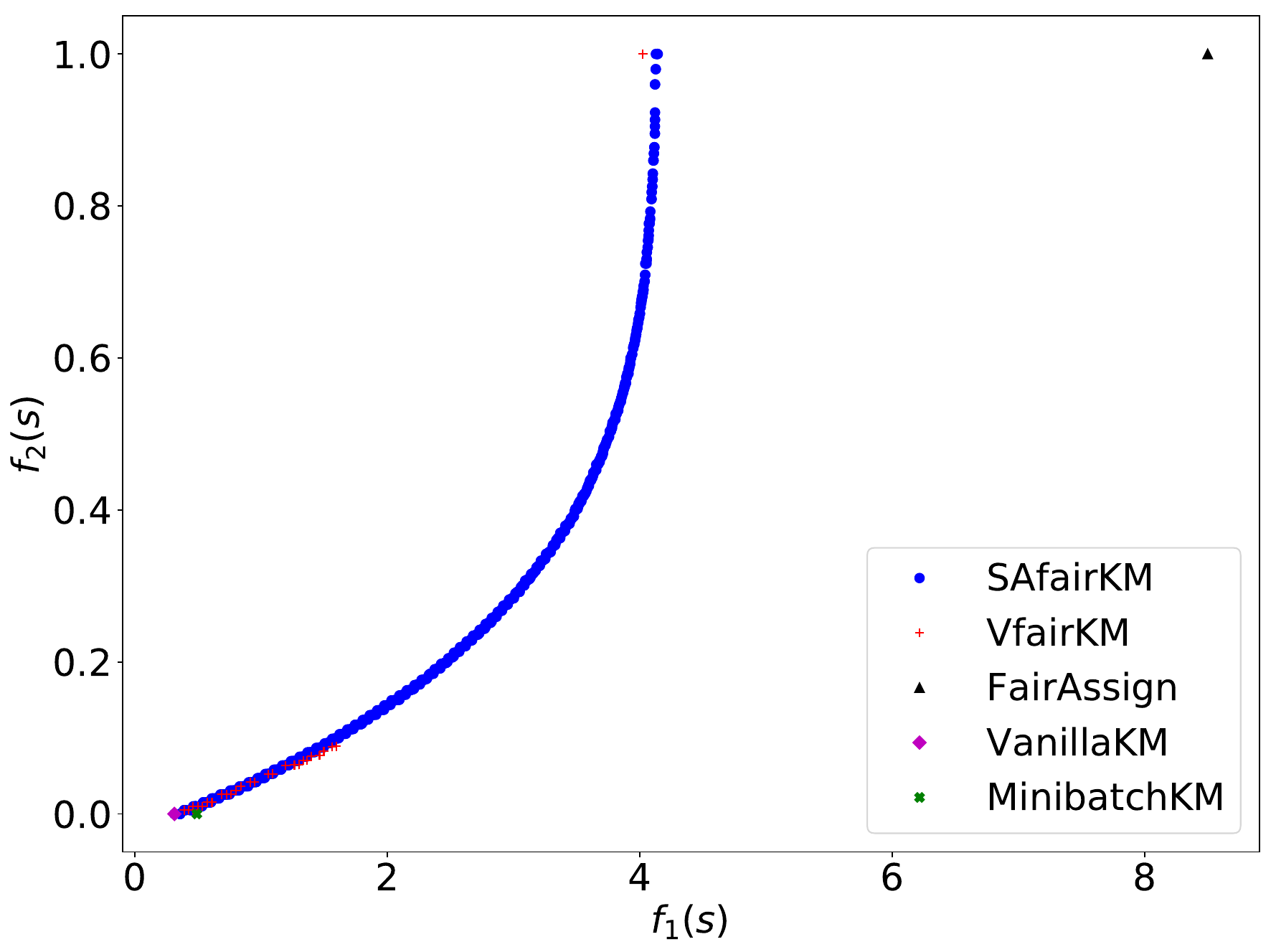}}
    \subfloat[Clust.~balance $b = 0$.]{\includegraphics[width = 0.25\linewidth]{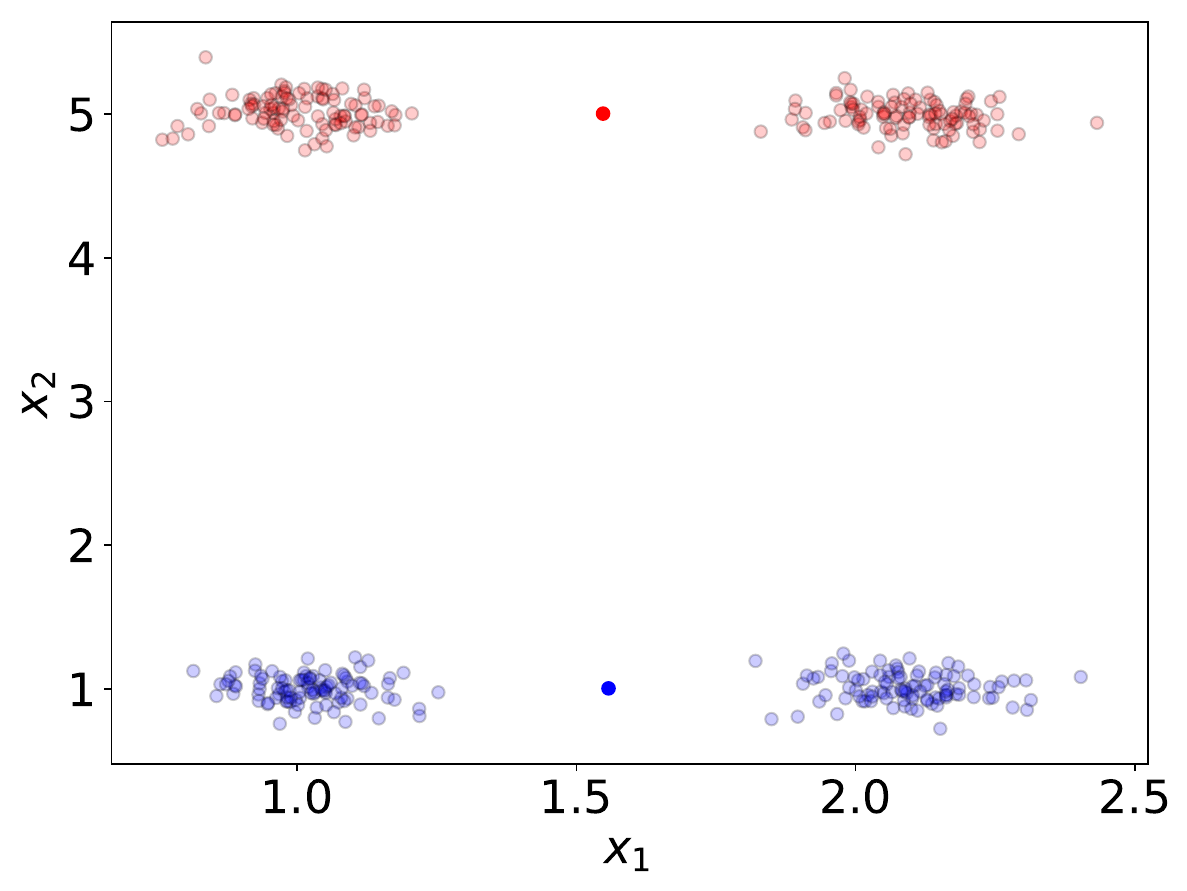}}
    \subfloat[Clust.~balance $b = 0.5$.]{\includegraphics[width = 0.25\linewidth]{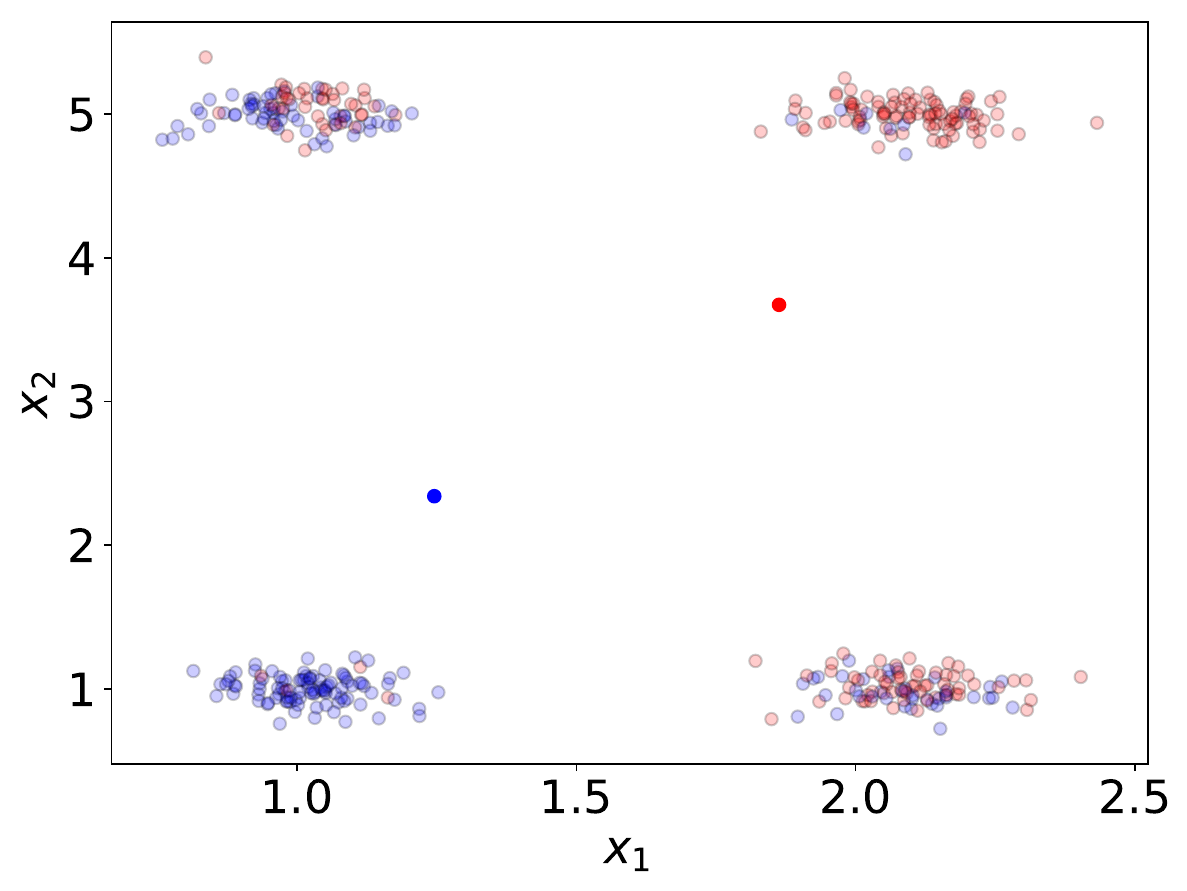}}
    \subfloat[Clust.~balance $b = 1$.]{\includegraphics[width = 0.25\linewidth]{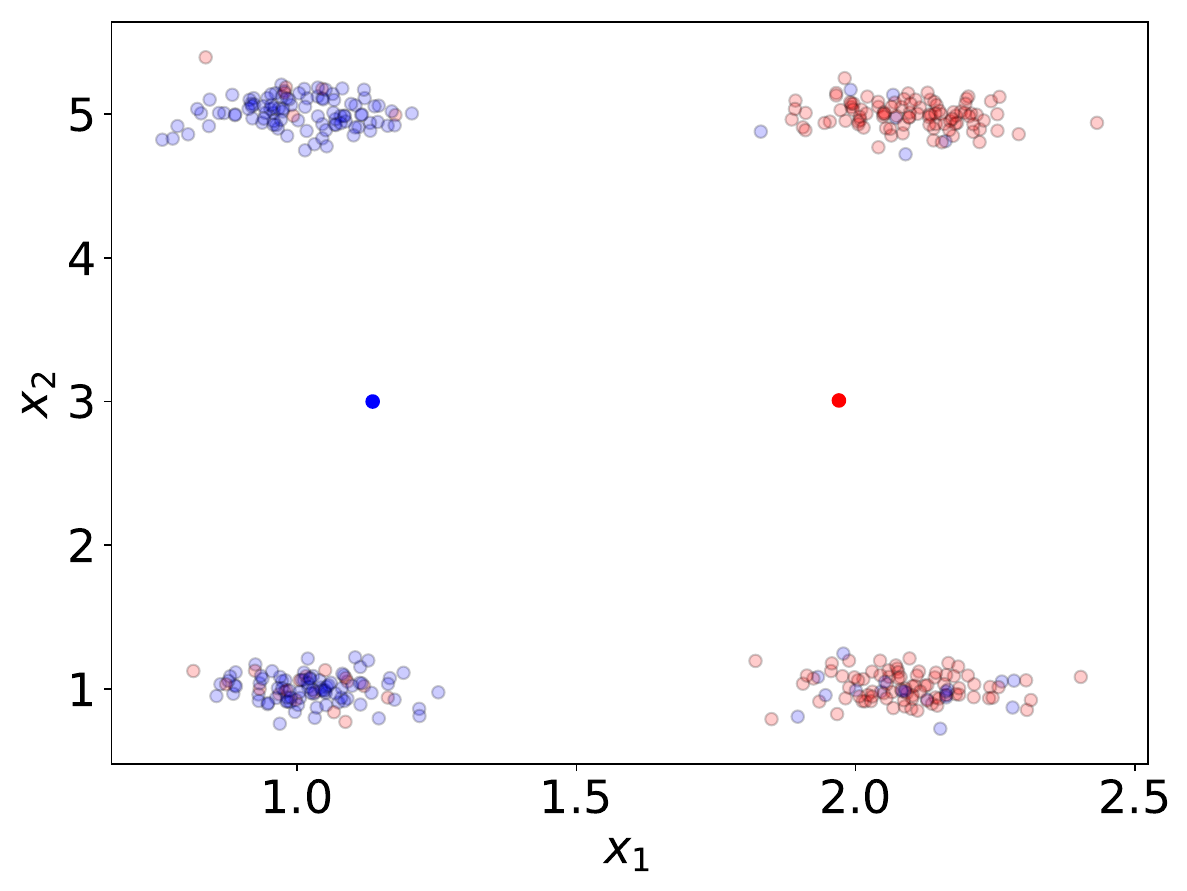}}
    \vspace{-2ex}
    \caption{Syn\_equal\_ds2 data: SAfairKM: $400$ iterations, $10$ starting labels, and $3$ pairs of $(n_a, n_b)$; VfairKM: $\mu_{\max} = 60$.\label{synDs2_equal}}
\vspace{-2ex}
\end{figure} 

\begin{figure}[!htb]
\setcounter{subfigure}{0}
    \centering
    \subfloat[Pareto front.]{\includegraphics[width = 0.25\linewidth]{Figures/SynData/Syn_unequal/Comparison_Pareto_front_dataSynthetic-unequal_numpts489_morecomp.pdf}}
    \subfloat[Clust.~balance $b = 0$.]{\includegraphics[width = 0.25\linewidth]{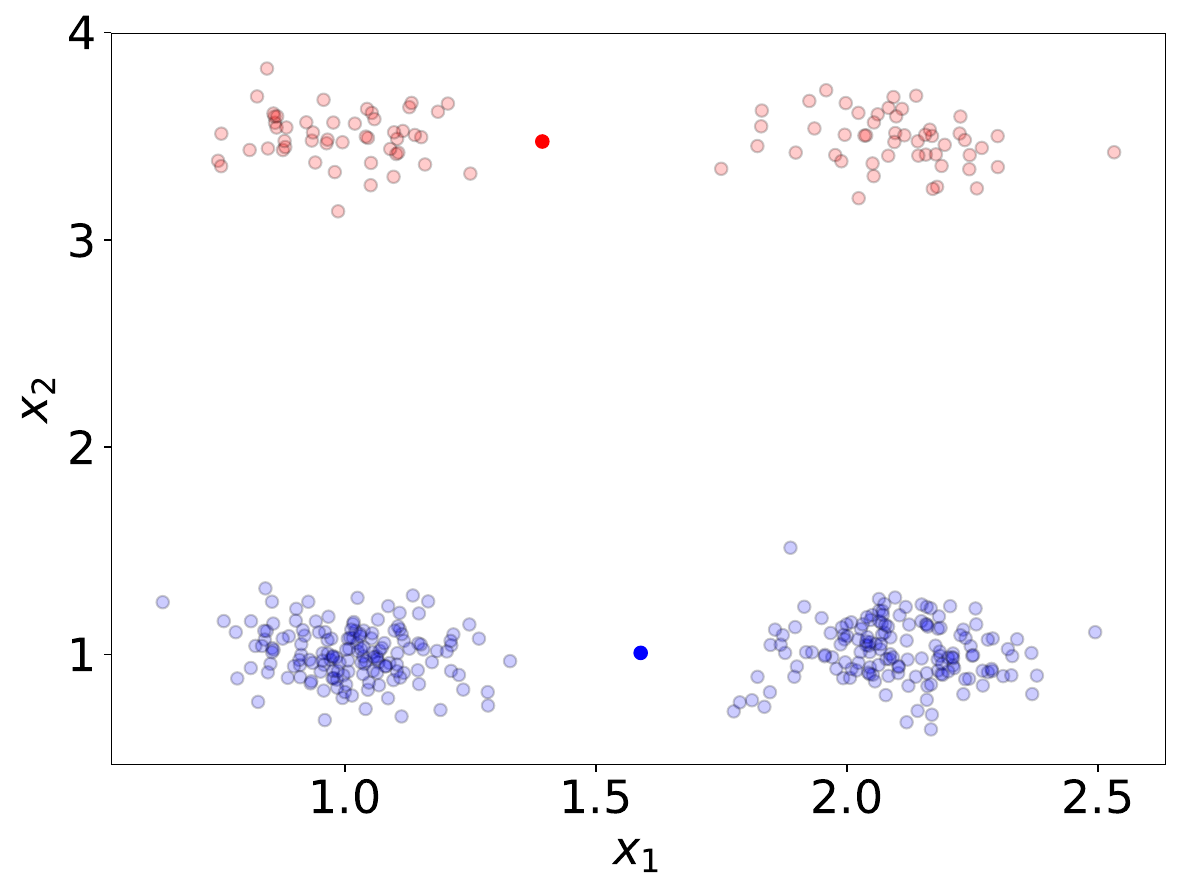}}
    \subfloat[Clust.~balance $b = 0.17$.]{\includegraphics[width = 0.25\linewidth]{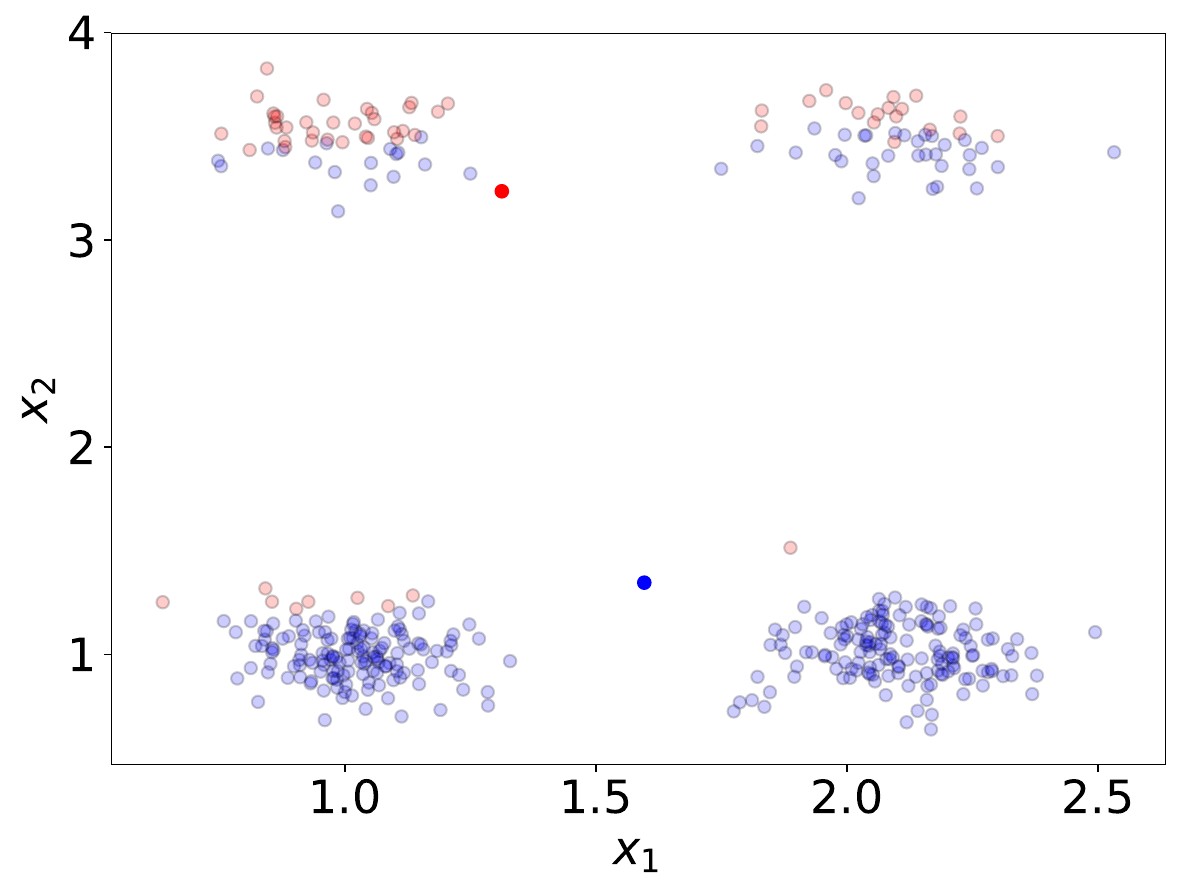}}
    \subfloat[Clust.~balance $b = 0.33$.]{\includegraphics[width = 0.25\linewidth]{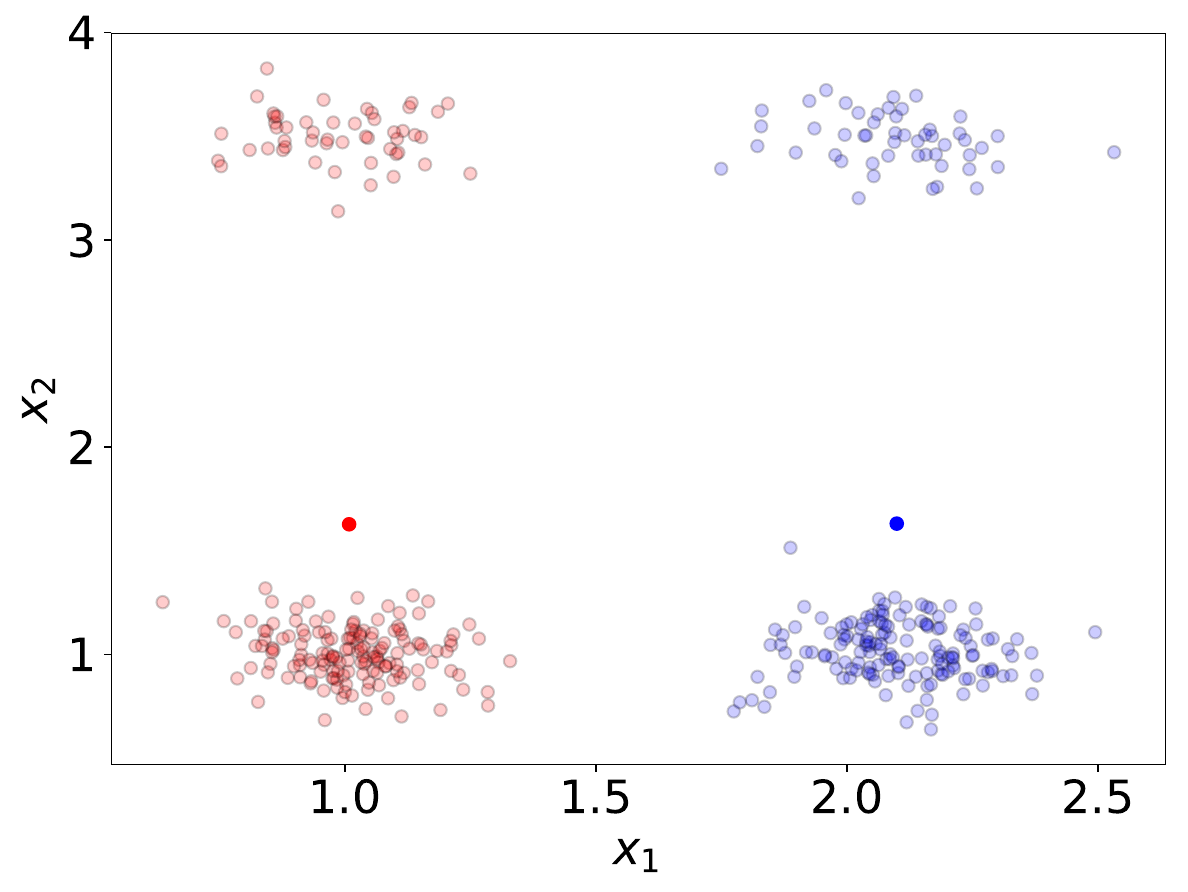}}
    \vspace{-2ex}
    \caption{Syn\_unequal\_ds2 data: SAfairKM: $400$ iterations, $10$ starting labels, and $3$ pairs of $(n_a, n_b)$; VfairKM: $\mu_{\max} = 0$. \label{synDs2_unequal}}
    \vspace{-2ex}
\end{figure}


\begin{figure}[htb]
\setcounter{subfigure}{0}
    \centering
    \subfloat[Adult.]{\includegraphics[width = 0.32\linewidth]{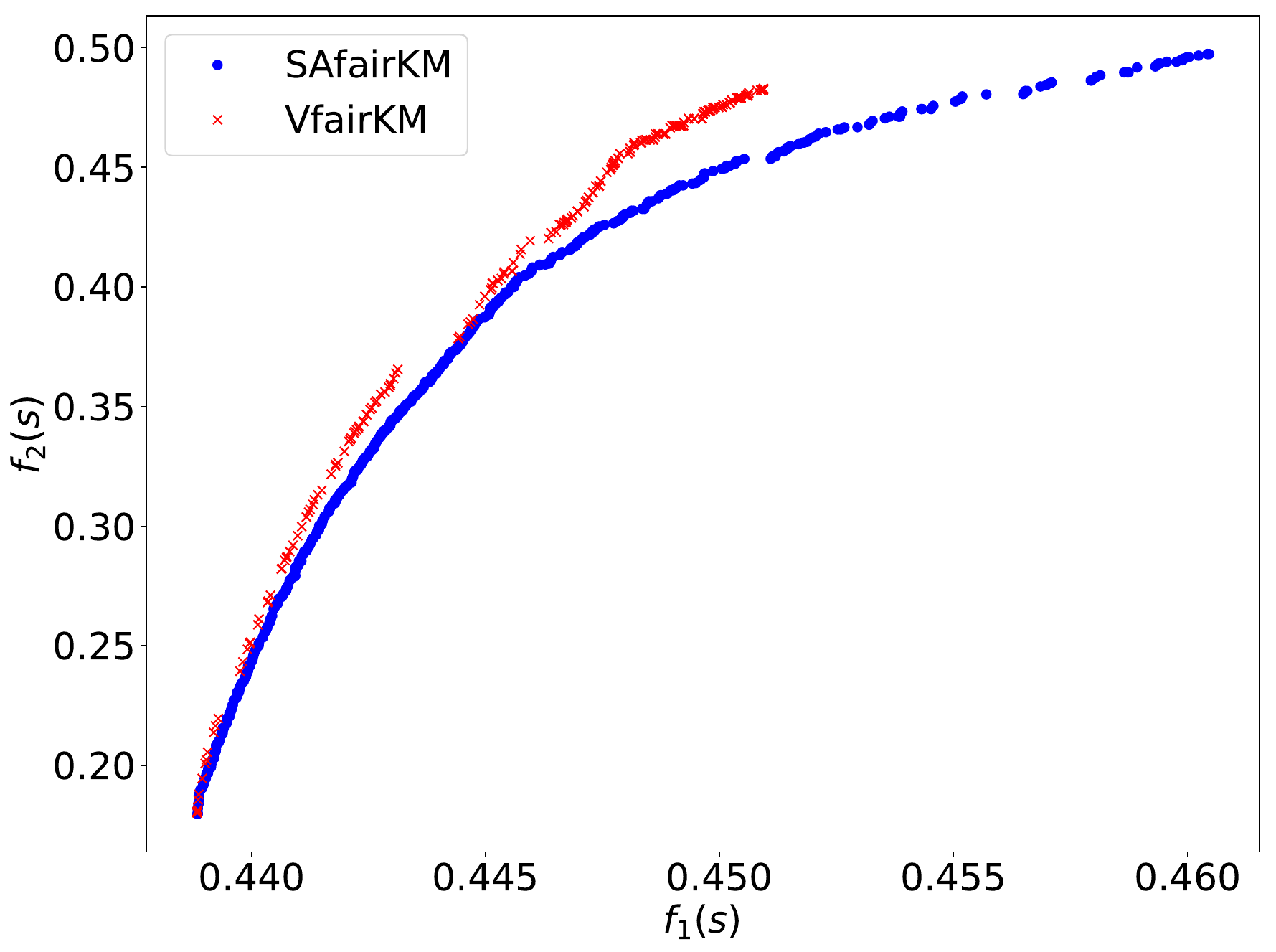}}
    \subfloat[Bank.]{\includegraphics[width = 0.32\linewidth]{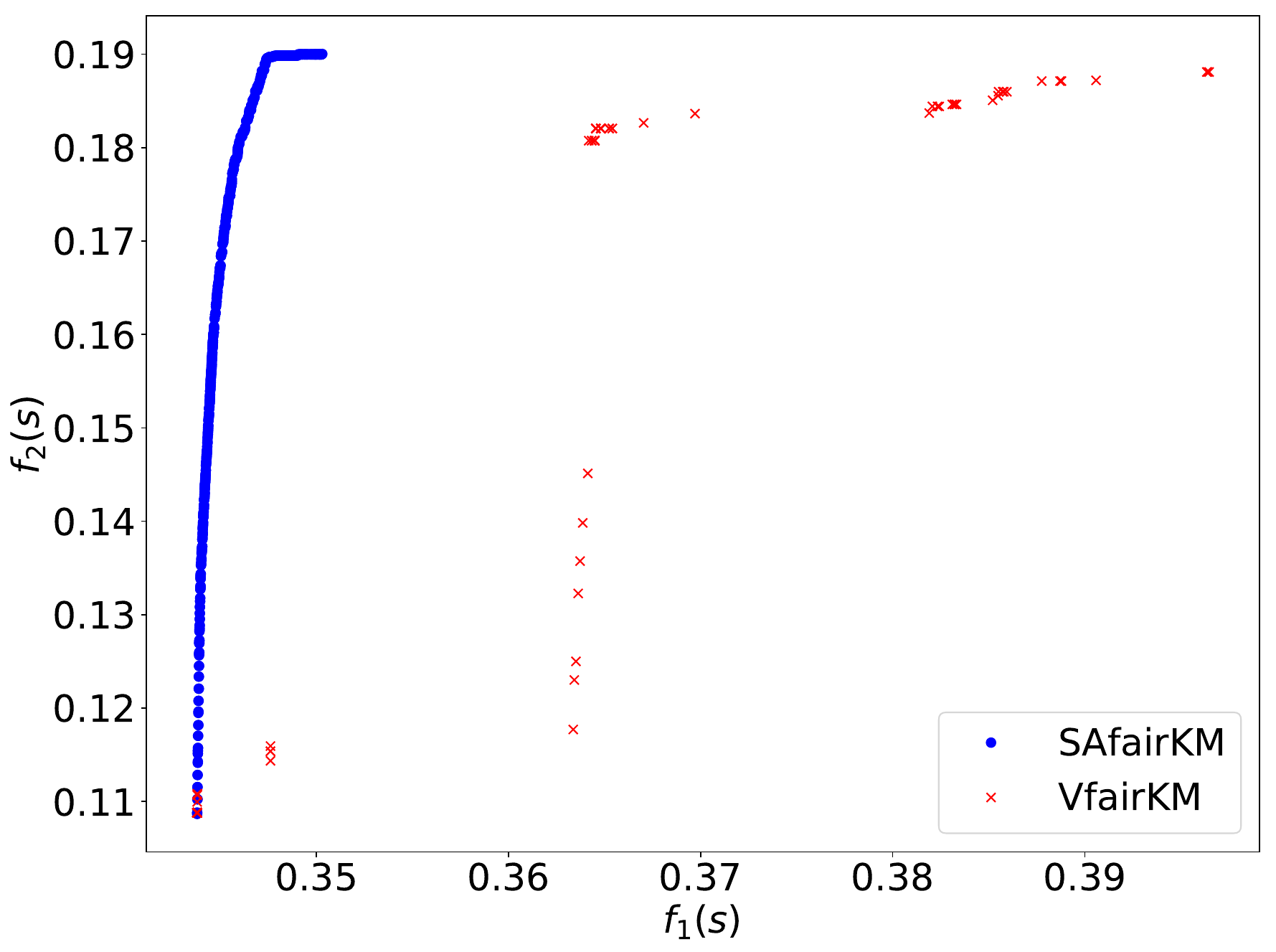}}
    \vspace{-2ex}
    \caption{Pareto fronts for $K = 5$: SAfairKM: $2500$ iterations for Adult and $1500$ iterations for Bank, $30$ starting labels, and $4$ pairs of $(n_a, n_b)$; VfairKM: $\mu_{\max} = 6190$ for Adult and $\mu_{\max} = 4790$ for Bank.\label{realDS_k5_seed1}}
\end{figure} 

\bibliographystyle{splncs04}
\bibliography{SAfairKM}

\end{document}